\documentclass[lettersize,journal]{IEEEtran}

\usepackage{amsmath,amsfonts}
\usepackage{array}
\usepackage{subcaption}
\usepackage{textcomp}

\usepackage{stfloats}
\usepackage{url}
\usepackage{verbatim}
\usepackage{graphicx}
\usepackage{cite}
\usepackage[ruled,linesnumbered]{algorithm2e}
\usepackage{subcaption}
\usepackage{ragged2e}
\usepackage[utf8]{inputenc}
\usepackage{tikz, pgfplots}
\usetikzlibrary{positioning}
\usepackage{amsmath}
\usepackage{breqn}
\usepackage{multirow}
\usepackage{multicol}
\usepackage{acronym}

\begin{document}
\title{Human-mediated Large Language Models for Robotic Intervention in Children with Autism Spectrum Disorders}

\author{Ruchik Mishra
, Karla Conn Welch, and Dan O Popa
\thanks{Ruchik Mishra is with graduate school of Electrical and Computer Engineering, University of Louisville, Kentucky, USA.}
\thanks{Karla Conn Welch is with the Faculty of
Electrical and Computer Engineering, University of Louisville, Kentucky, USA.}

\thanks{Dan O Popa is with the Faculty of
Electrical and Computer Engineering, University of Louisville, Kentucky, USA.
This project was supported by the National Institutes of Health (NIH) and the National Science Foundation (NSF) through a Smart and Connected Health (SCH) grant \#1838808 and in part by NSF through grant EPSCoR-OIA \#1849213.}
}




\maketitle

\begin{abstract} 
The robotic intervention for individuals with Autism Spectrum Disorder (ASD) has generally used pre-defined scripts to deliver verbal content during one-to-one therapy sessions. This practice restricts the use of robots to limited, pre-mediated instructional curricula. In this paper, we increase robot autonomy in one such robotic intervention for children with ASD by implementing perspective-taking teaching. Our approach uses large language models (LLM) to generate verbal content as texts and then deliver it to the child via robotic speech. In the proposed pipeline, we teach perspective-taking through which our robot takes up three roles: initiator, prompter, and reinforcer. We adopted the GPT-2 $+$ BART pipelines to generate social situations, ask questions (as initiator), and give options (as prompter) when required. The robot encourages the child by giving positive reinforcement for correct answers (as a reinforcer). In addition to our technical contribution, we conducted ten-minute sessions with domain experts simulating an actual perspective teaching session, with the researcher acting as a child participant. These sessions validated our robotic intervention pipeline through surveys, including those from NASA TLX and GodSpeed. We used BERTScore to compare our GPT-2 $+$ BART pipeline with an all GPT-2 and found the performance of the former to be better. Based on the responses by the domain experts, the robot session demonstrated higher performance with no additional increase in mental or physical demand, temporal demand, effort, or frustration compared to a no-robot session. We also concluded that the domain experts perceived the robot as ideally safe, likable, and reliable.

\end{abstract}
\section{Introduction}\label{sec1}

\begin{figure*}[h!]
    \centering
    \subfloat[Study setup]{%
        \centering
        \includegraphics[scale=0.175]{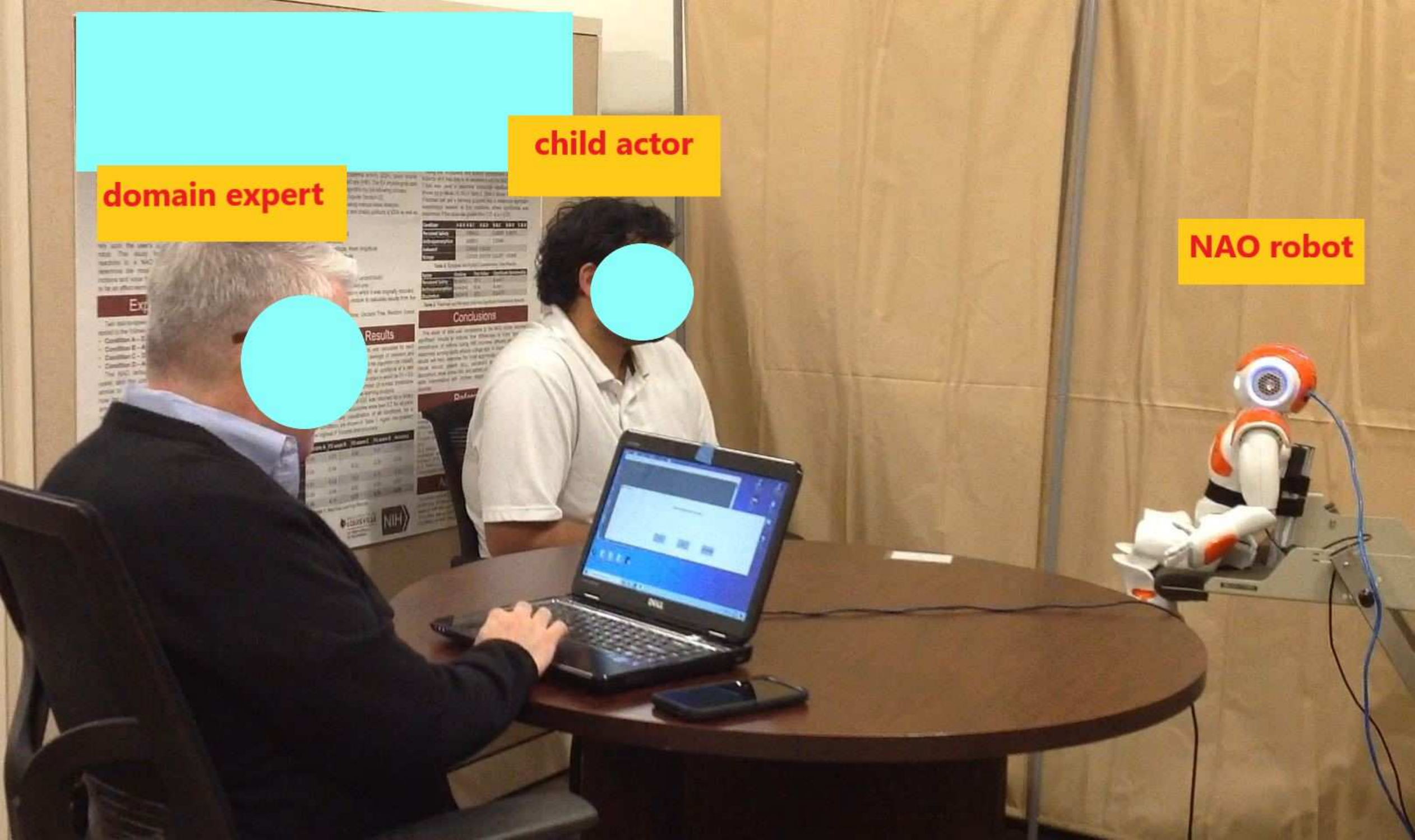}
        \label{fig:teaser_1}
    }
    \hfill
    \subfloat[NAO robot for teaching perspective-taking to children with ASD.]{%
        \centering
        \includegraphics[scale=0.50]{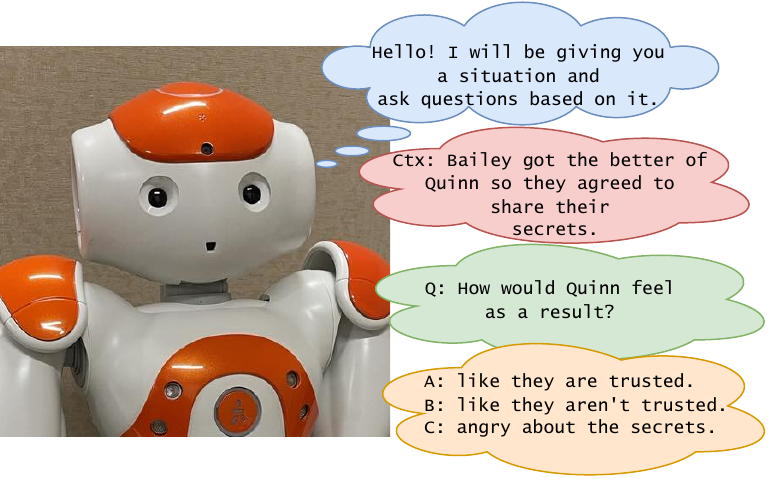}
        \label{fig:teaser_2}
    }
    \caption{The setup shows the domain expert (actual) and the child with ASD (actor) in a session with the NAO robot. Here `Ctx' refers to the context describing a situation, `Q' is the question based on that context and, A-C are the options consisting of a possible answer.}
    \label{fig:teaser}
\end{figure*}
Social robots have positively impacted people's lives in the past decade \cite{lambert2020systematic}. They currently share various scenarios of coexistence with humans. This coexistence accounts for Human-Robot Interaction (HRI) in domestic environments like the Roomba robot \cite{forlizzi2006service}, or of a more interactive robot like Autom that was built to make behaviour changes in a more naturalistic user setting \cite{kidd2008robots}.

Within healthcare, social robots have undertaken diverse roles too. These roles have required  the robot to be involved in a multi-modal interaction with humans. Examples include the telepresence robot discussed in \cite{tsui2011exploring}, facilitating both visual and verbal communication. Other examples of robots being used in healthcare include robots used in surgery \cite{okamura2004methods,el2020review}. Moreover, robots have found application in psychological interventions, representing another domain where their presence can be immensely notable, as has been seen in ASD interventions \cite{bartl2021robot}.
\cite{saral2022autism}. 
Autism Spectrum Disorder, as described by the DSM-5, describes individuals that exhibit: 1) challenges in social communication and interaction and 2) restricted and repetitive behaviors \cite{regier2013dsm}. Since there is no cure for ASD, there are countless interventions and treatments, as has been highlighted by the authors in \cite{saral2023autism}.
As robotic intervention of ASD is getting popular with evidence of positive outcomes \cite{steinbrenner2020evidence}, the development and evaluation of such platforms have gained momentum. Since the population of people with ASD is 1.5\% \cite{thabtah2020new}, with every 1 of 54 children in the US alone, sustainability in robotic intervention becomes essential.

Since children with ASD cannot understand social cues and situations, a robot that can teach such skills has been a widely proposed solution for both verbal and non-verbal communications \cite{ntaountaki2019robotics, wijayasinghe2016human}. So far, the solutions in the literature for robotic intervention have had limited scope for scalability because of the number of finite responses, actions, or pre-defined contents \cite{scassellati2018improving} that the robot can generate. Hence, there is a need to bridge this gap between the deep learning literature that can be used to generate content (speech, motions, etc.) that the robot can deliver specifically for children with ASD to help them understand various social skills.

To the best of our knowledge, this work is novel in its approach to creating an embodied speech AI agent that can independently generate verbal content during a perspective-taking teaching session for children with ASD. This introduction of autonomy in classical human-led interventions or even teleoperated robot-led interventions shows the potential for more general interventions aimed at individuals with ASD. We compared our GPT-2 $+$ BART pipeline to an all GPT-2 pipeline and found the former to be better for question, option, and overall generation of both questions and options (explained more in Section \ref{results}). In addition to this technical contribution, we also conducted ten-minute sessions with domain experts and a child actor, with the robot acting as an initiator, prompter, and reinforcer. Based on the survey data collected at the end of each of these ten-minute sessions, we found that the robot session demonstrated higher performance with no additional increase in mental demand, physical demand, temporal demand, effort, or frustration as compared to a no-robot session (NASA TLX data). We also concluded that the domain experts perceived the robot as ideally safe, likable (GodSpeed data), and the entire session to be technically relevant and reliable (Appropriateness). 

This paper has been arranged in the following way: Section \ref{Related_work} outlines the existing work in the literature related to robots in ASD intervention (Section \ref{ASD_inter}), robots to teach perspective-taking (\ref{perspective_teach}), and text generation (Section \ref{content_gen}). Section \ref{methodology} which describes the workflow highlighting how the robot acts as an initiator, prompter and reinforcer. Further, Section \ref{problem_formula} describes the problem formulation that we use in this paper. Section \ref{results} outlines the results and discussions, followed by Section \ref{limit_future}, which describes the limitations and scope for future work. Finally, Section \ref{conclusion} describes the conclusion.

\section{Related Work}\label{Related_work}
\subsection{Robots in ASD intervention}\label{ASD_inter}
The robotic intervention of children with ASD has witnessed variety in both the choice of robot and the respective therapeutic or educational setting. This approach has facilitated the integration of classical methodologies combined with a socially assistive robot, which has motivated researchers to examine its effectiveness. In this direction, the authors in \cite{begum2015measuring} have evaluated a Wizard-of-Oz style therapy following the Applied Behavior Analysis (ABA) structure. Based on the efficacy metrics proposed (skill execution and prompt dependency), they could make better conclusions on the clinical merits of a robot-assisted intervention of ASD, which contradicts the HRI metrics that considers gaze, communication, and affect. They also demonstrated the positive effects of such an intervention on one of their study participants. Similar studies done by the authors in \cite{cao2019robot} have also contributed to positive conclusions about social robots.
 
The authors in  \cite{van2020adherence} have tested their hypothesis of positive outcomes in a Pivotal Response Treatment (PRT) with robot assistance. They reported that robotic assistance in PRT had a positive effect at the end of the session and showed likeability towards the robot. Furthermore, the results obtained in \cite{takata2023social} pointed towards the generalization of improvement in the sociability of the participants in non-robot scenarios. In \cite{boccanfuso2017low}, a new low-cost platform was field tested to see the efficacy of a robot-assisted intervention to improve spontaneous speech, communication, and social skills. Based on statistical analysis, the authors could compare between a robot- and no-robot-assisted groups and could claim relatively higher growth in the former scenario. In addition to research articles, several review articles have pointed towards the productivity of using robot-assisted interventions for individuals with ASD \cite{pennisi2016autism, stolarz2022personalized, cabibihan2013robots}. 


Although both the robot-led and human-led sessions have proven beneficial to fetch positive outcomes, robots, compared to humans, have yet to be considered serious agents owing to the lack of sophistication they can offer \cite{salimi2021social, holeva2022effectiveness}. 
Although in \cite{huijnen2019roles}, the authors used the KASPER robot to underline the roles it can take up in a therapy session, it still has limitations in the finite reaction possibilities the robot can generate. The robot's finite response generation ability also adds to the effort a therapist has to put in during a therapy session. As pointed out in  \cite{scassellati2012robots}, the therapist's involvement in generating content or controlling the robot, as in a Wizard-of-Oz approach, is not sustainable in the long run. Hence, a more sustainable autonomous/semi-autonomous solution is needed in which the robot can take up the role of an initiator that can generate and deliver content during a therapy session \cite{huijnen2019roles, scassellati2012robots}.

\begin{figure*}[h!]
    \centering
        \includegraphics[scale = 0.65]{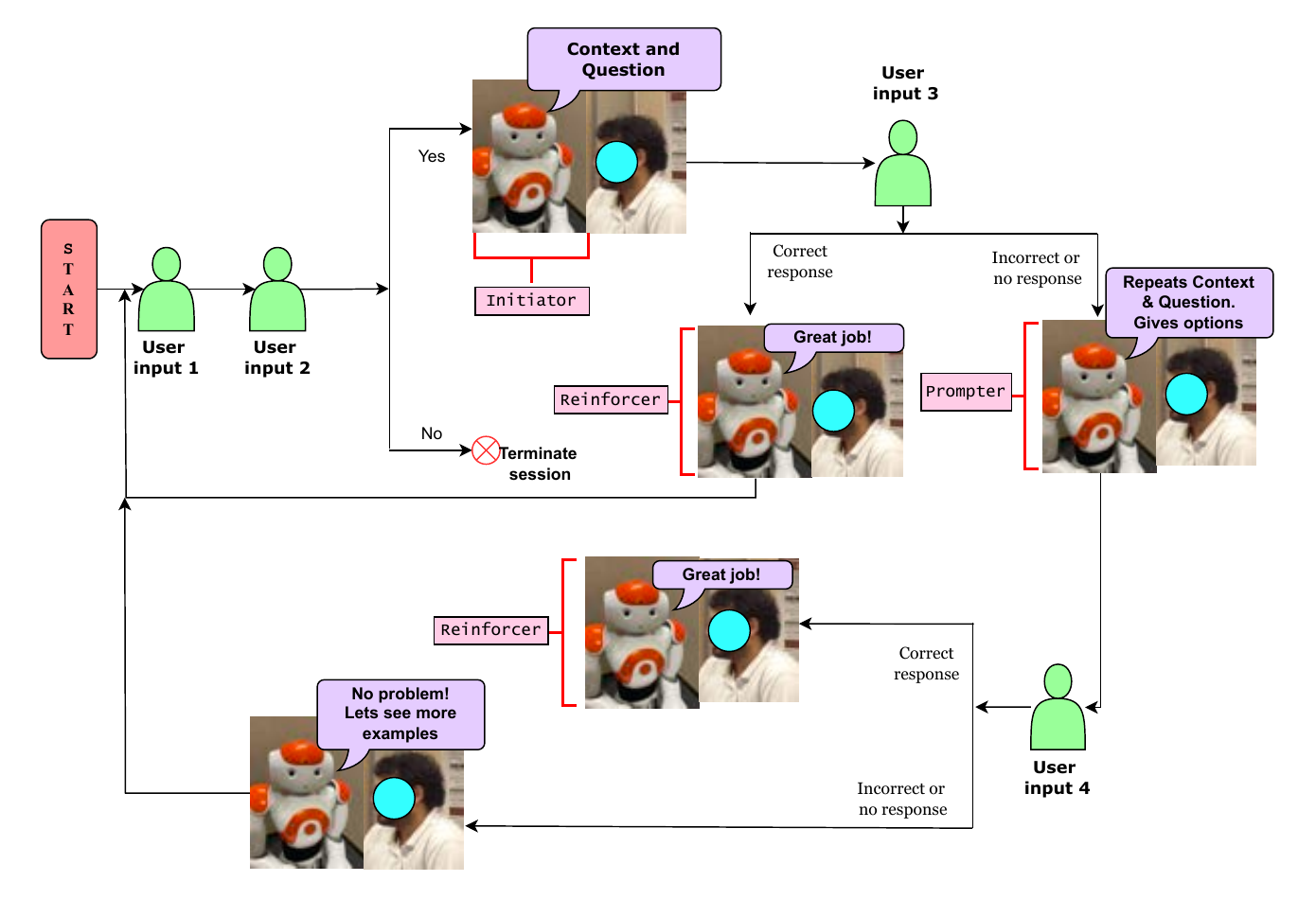}
        \caption{Pipeline of the human-mediated autonomous system.}
    \label{fig:pipeline}
\end{figure*}
\subsection{Robots to teach perspective-taking and theory of mind}\label{perspective_teach}
Theory of mind (ToM), as defined in \cite{baron1985does}, is the ability to understand another person's mental state to aid the understanding of their social scenario (perspective-taking). Unlike neurotypical children, children with ASD have difficulties perceiving how another person might react in a particular social situation or what personal thoughts might arise out of them in such scenarios \cite{gould2011teaching, welsh2019teaching, kimhi2014theory}. In order to enhance the understanding of these social scenarios, many approaches have been adopted in the ToM interventions.

The authors in \cite{hoddenbach2012individual} have employed a combination of tasks, including learning different scenarios through visual and verbal modalities, to understand perspective-taking. Similarly, in \cite{begeer2015effects}, verbal communication and graphical cues-based approach have been followed to assess the beliefs of an individual evaluated based on social scenarios.

Human therapists were involved throughout the session for both of the above ToM-based interventions. Unlike these approaches, the authors in \cite{rudovic2018personalized, scassellati2018improving} have used a social robot to teach social skills aligned towards the ToM and perspective-taking realm. However, in these studies, the robots followed a pre-defined script of verbal communication to deliver the contents.

\subsection{Content generation using Language Models for ASD intervention}\label{content_gen}
Text generation, as such, is not a new problem in the natural language processing (NLP) literature. Recently, language models have seen a boom after the introduction of the transformers \cite{vaswani2017attention} architecture on various NLP tasks. They have outperformed their RNN-based counterparts for machine translation, text generation, etc. \cite{ezen2020comparison, rahali2023deeppress} because RNNs suffer from the vanishing gradient problems because of their sequential nature \cite{le2016quantifying,hochreiter1998vanishing}. On the other hand, transformers use a self-attention mechanism, which provides a more parallel computational scheme that is more effective in generating long textual sequences \cite{bahdanau2014neural}. 

There can be three broad categories to classify Large Language Models (LLMs) (based on the transformers architecture) for text generation. These include the encoder-only architectures like Bidirectional Encoder Representations from Transformers (BERT) \cite{devlin2018bert} where beam search can be used to collect the most probable subsequent tokens to predict in a sequential manner \cite{chan-fan-2019-bert,chan2019recurrent}. The second type of model is the decoder-only architectures like OpenAI's Generative Pre-trained Transformer (GPT), which has immensely been used for text generation across different fields of study \cite{luo2022biogpt, chintagunta2021medically, li2022tod4ir}. The third category comprises a model with an encoder-decoder architecture like the BART model, which is used for tasks like text summarization, classification, etc. \cite{lewis2020bart}.
\begin{figure*}[h!]
    \centering
        \includegraphics[scale = 0.6]{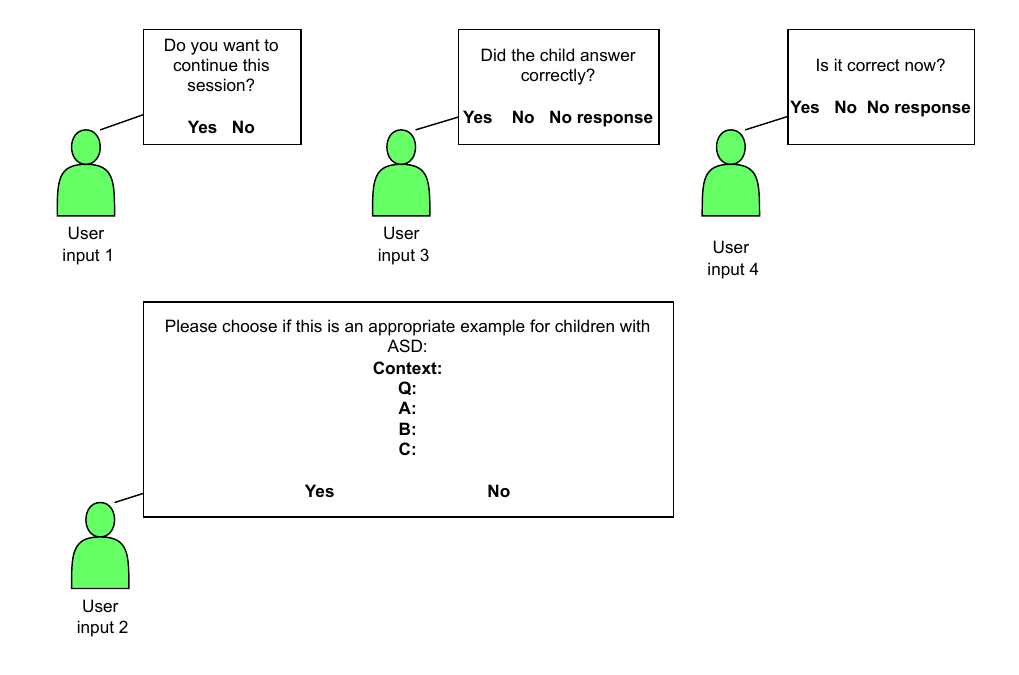}
        \caption{User inputs that the domain experts are asked to give in on the computer based on the questions asked on the GUI.}
    \label{fig:pipeline}
\end{figure*}
\begin{figure*}[h!]
    \centering
        \includegraphics[scale = 0.50]{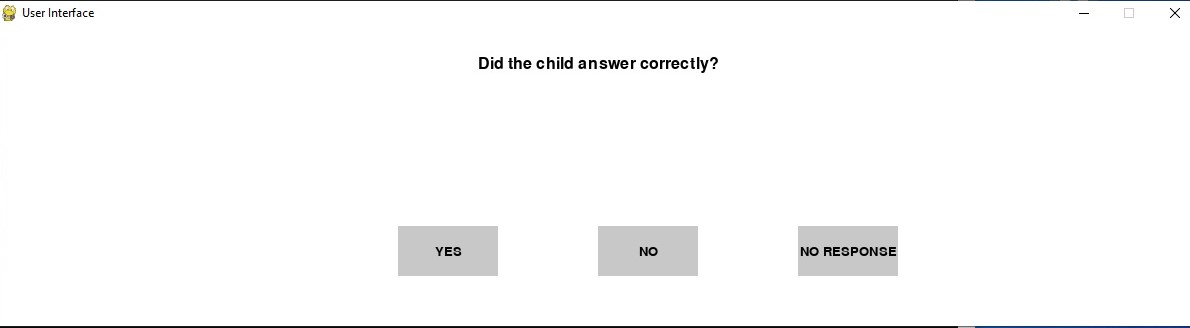}
        \caption{Example of how the actual GUI looks like for user input 3.}
    \label{fig:gui}
\end{figure*}
Recently, LLMs have been used for robotic-specific tasks \cite{ye2023improved, huang2023grounded, vemprala2023chatgpt}. This also involves HRI scenarios where goal-oriented LLM usage has been embodied as a robot. Examples include robots being used as a teaching assistant \cite{lekova2022making}, involved in task planning \cite{xie2023chatgpt} or, being used as an intervention tool in healthcare \cite{bertacchini2023social}. 

Further, generating content using LLMs for a robotic intervention of ASD is still an active area of research. To the best of our knowledge, the only work that combines a social robot with using LLMs for ASD intervention is \cite{bertacchini2023social}. In this, the authors have used a Pepper robot integrating ChatGPT to meet the therapeutic goals of individuals with ASD. However, the structured component used for the problem-solving session is scripted, and the LLM integration is essentially an add-on to provide additional feedback and insights. Hence, it still does not fully meet the autonomy requirements for content generation during therapy sessions, which the authors pointed out in \cite{huijnen2019roles}. 

Hence, this paper is motivated by the absence of a system that can be used for real-time content generation and take another step towards delivering high-end therapeutic scenarios in a goal-oriented therapy.

\begin{figure*}
    \centering
    \includegraphics[scale = 0.60]{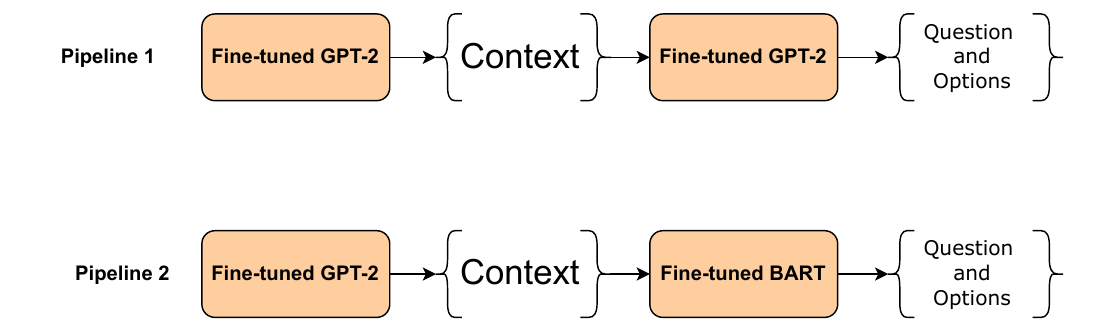}
    \caption{LLM pipelines used and compared in this paper.}
    \label{fig:model_pipeline}
\end{figure*}

\section{Methodology}\label{methodology}

\subsection{Human mediated autonomy}\label{human_mediated_aut}
In the pipeline shown in Figure \ref{fig:pipeline}, the robot takes up three roles: initiator, prompter, and reinforcer similar to the definitions suggested by the authors in \cite{huijnen2019roles}. The session starts with user input 1, where the domain expert is asked through a graphical user interface (GUI) on the computer if they want to continue the session. If they select yes, they are then shown the context (which describes a social situation), question and the respective options. If they feel that this set of context, question and, options are appropriate for children with ASD, then they press yes. This allows the robot to take the role of the initiator, where it speaks the context and the question only and waits for a response from the child.  

This generation of context and the question is not pre-determined. Rather, it is generated by our LLM model (described in section \ref{problem_formula}). If the child answers the question correctly (decided by the user input 3), the robot acts as a reinforcer and provides positive reinforcement by praising them on a job well done. However, if the child does not answer or answers incorrectly (decided by the user input 3), the robot takes up the role of the prompter. 

In the role of a prompter, the robot repeats the context and the question and now speaks out the options to choose from. If the child answers correctly (decided by the user input 4), the robot gives positive reinforcement at this stage by congratulating them on correctly answering. On the other hand, if the user input 4 is either `no' or `no response', the robot returns to user input 1 and repeats this cycle.  

As described above, this is a semi-autonomous system, meaning the autonomy lies in the content generation part. However, the decisions on whether the content is appropriate for children with ASD and whether the child answers correctly are made by the domain expert through a computer. This establishes control over what the robot can do, which is essential given a population such as children with ASD. 

\subsection{Autonomous content generation using LLMs}
As shown in Figure \ref{fig:model_pipeline}, we use two LLM pipelines to generate the questions and the options. However, the context generation through fine-tuned GPT-2 remains the same for both pipelines. The problem formulation is explained in Section \ref{problem_formula}.

For fine-tuning both of our LLM pipelines, we use the SOCIALIQA dataset, which is a social commonsense reasoning benchmark \cite{sap2019social}. This dataset has 38,000 data points containing contexts, questions based on those contexts, and multiple-choice options for those questions. The questions broadly fall into three categories. Based on the context, the questions can inquire 1) about motivation, 2) about next steps, or 3) emotional reactions.

\begin{figure*}[h!]
\centering
\resizebox{0.95\textwidth}{!}{
\begin{tikzpicture}[
SIR/.style={rectangle, draw=red!60, fill=red!5, very thick,  minimum width=14mm, minimum height=5mm},
BART/.style={rectangle, draw=blue!60, fill=blue!5, very thick, minimum width=33mm, minimum height=8mm},
mask/.style={rectangle, draw=green!60, fill=green!5, very thick, minimum width=22mm, minimum height=8mm},
loss/.style={rectangle, draw=black!60,rounded corners=8pt,fill=black!5, very thick, minimum width=22mm, minimum height=8mm}, 
summation/.style={circle, draw=black!60, fill=black!5, very thick, minimum size = 10mm}, 
]

\node[loss](cross_ent1)[]{Cross entropy};

\node[BART](BART)[below=0.5cm of cross_ent1]{BART};
\node[] (Context) [below=1em of BART] {$\textrm{ctx}_{i} + <\textrm{mask}>_{q_{i}\textrm{[} \textrm{:}j \textrm{]}}$};
\draw[->, very thick] (Context.north)  to node[right] {} (BART.south);
\node[BART](BART2)[right=1.2cm of BART]{BART};
\node[] (Context2) [below=1em of BART2] {$\textrm{ctx}_{i} + \mathcal{Q}_{i} + <\textrm{mask}>_{a_{i}\textrm{[} \textrm{:}k \textrm{]}}$};
\draw[->, very thick] (Context2.north)  to node[right] {} (BART2.south);
\node[BART](BART3)[right=1.2cm of BART2]{BART};
\node[] (Context3) [below=1em of BART3] {$\textrm{ctx}_{i} + \mathcal{Q}_{i} + \mathcal{A}_{i} + <\textrm{mask}>_{b_{i}\textrm{[} \textrm{:}p \textrm{]}}$};
\draw[->, very thick] (Context3.north)  to node[right] {} (BART3.south);



\node[BART](BARTn)[right=1.2cm of BART3]{BART};
\node[text width=3cm] (Context4) [below=1em of BARTn] {$\textrm{ctx}_{i} +$ $\mathcal{Q}_{i} +$ $\mathcal{A}_{i} +$ $\mathcal{B}_{i} +$ $<\textrm{mask}>_{c_{i}\textrm{[} \textrm{:}r \textrm{]}}$};
\draw[->, very thick] (Context4.north)  to node[right] {} (BARTn.south);

\node[loss](cross_ent2)[above=0.5cm of BART2]{Cross entropy};
\node[loss](cross_ent3)[above=0.5cm of BART3]{Cross entropy};
\node[loss](cross_ent4)[above=0.5cm of BARTn]{Cross entropy};

\draw[->, very thick] (BART.north)  to node[right] {} (cross_ent1.south);
\draw[->, very thick] (BART2.north)  to node[right] {} (cross_ent2.south);
\draw[->, very thick] (BART3.north)  to node[right] {} (cross_ent3.south);
\draw[->, very thick] (BARTn.north)  to node[right] {} (cross_ent4.south);

\node[summation](summation)[above=3cm of BART2, xshift = 2.2cm]{$\sum$};
\draw[->, very thick] (cross_ent1.north)  to node[right,xshift=-6.5em] {$\mathcal{L}_{q_{i}}$} (summation.south west);

\draw[->, very thick] (cross_ent2.north)  to node[right,xshift=0em] {$\mathcal{L}_{a_{i}}$} ([xshift=-0.4em]summation.south);

\draw[->, very thick] (cross_ent3.north)  to node[right,xshift=-1.9em] {$\mathcal{L}_{b_{i}}$}([xshift=0.4em]summation.south);

\draw[->, very thick] (cross_ent4.north)  to node[right,xshift=4.0em] {$\mathcal{L}_{c_{i}}$}
([xshift=0em]summation.south east);

\node[](total_loss)[above=4mm of summation]{$\mathcal{L}$};
\draw[->, very thick] (summation.north)  to node[right] {} (total_loss.south);

\end{tikzpicture}
}
    \caption{BART pipeline used in this paper. $\textrm{ctx}_{i}$ is the context for the $i^{th}$ data point. Similarly, $<\textrm{mask}>_{q_{i}\textrm{[} \textrm{:}k \textrm{]}}$ denotes the total number of tokens in the questions}
    \label{fig:model_add_loss}
\end{figure*}
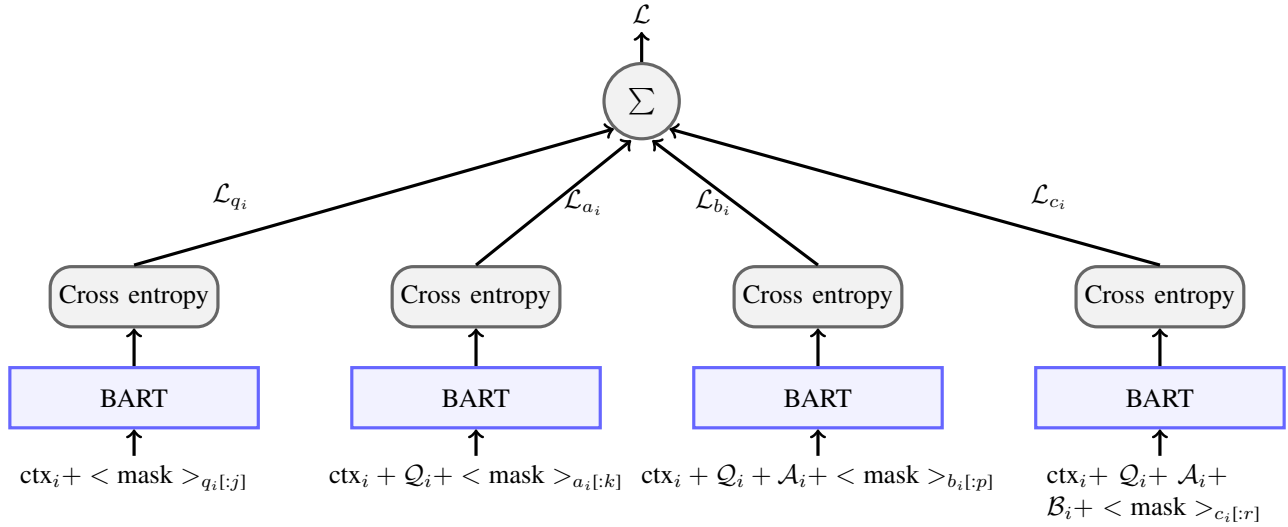

\subsection{Data collection}
In this study, we not only make a technical contribution by leveraging LLMs to introduce autonomy in perspective-taking sessions using a robot but also collect the opinions of domain experts on some surveys. We asked four domain experts to perform a 10-minute session each (independently) with the robot and a participant (not an actual child with ASD but a child actor). The domain experts included one psychologist, one neurophysician, one professor from the special education domain, and a clinical assistant professor from the special education domain, each having substantial experience conducting sessions in therapeutic, clinical, and educational settings with children with ASD. At the end of the 10-minute session, we asked them to fill out the NASA Task Load Index (NASA TLX), comparing their session with the robot with classical non-robotic sessions for perspective-taking. In addition, we asked them to fill out the Godspeed survey to understand their perceptions of the robot. Lastly, we prepared an additional survey (called the Appropriateness survey) specific to our session to understand how useful our human-mediated autonomous robotic pipeline was for teaching perspective-taking to children with ASD.

\section{Problem Formulation}\label{problem_formula}

\begin{figure*}[h!]
    \centering
    \subfloat[T-SNE plot for 2 components.]{%
        \centering
        \includegraphics[scale=0.4]{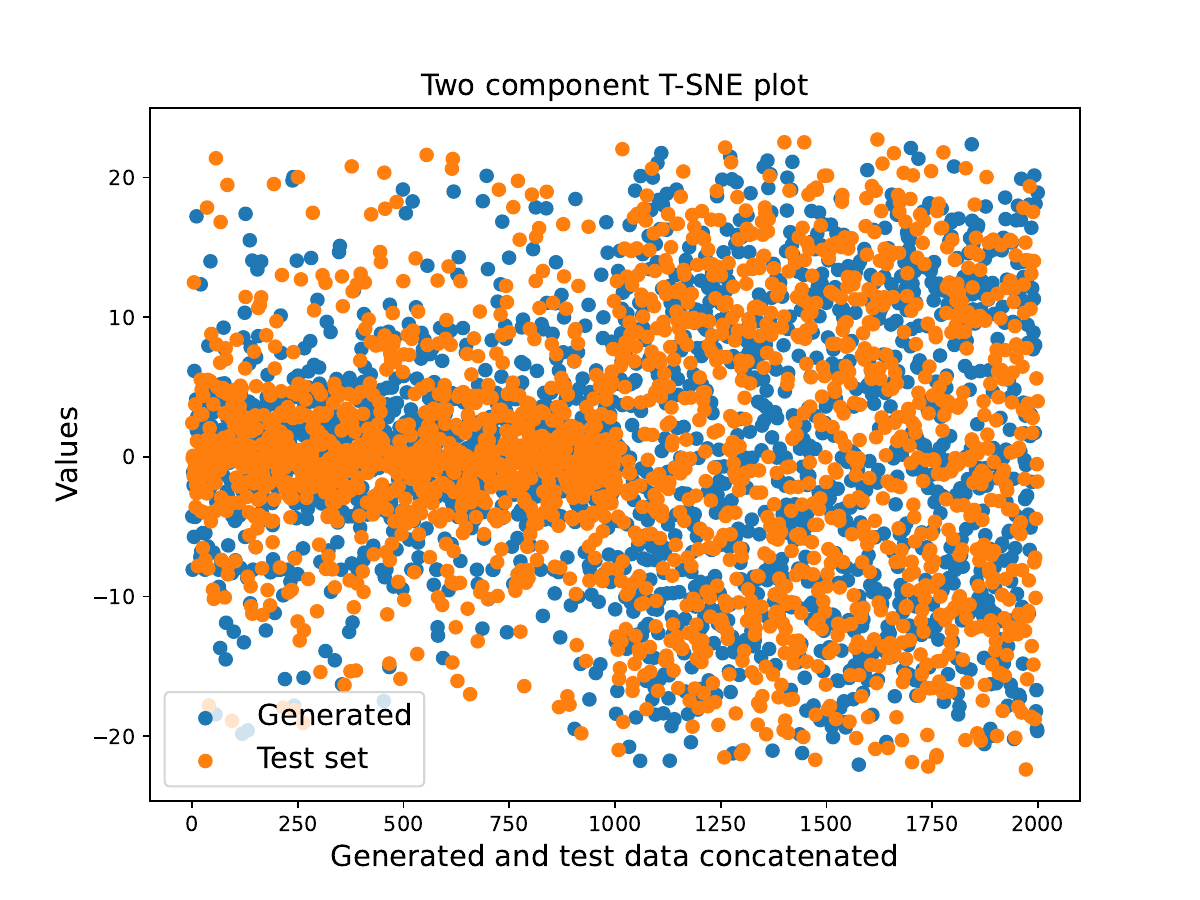}
        \label{fig:tsne}
    }
    \hfill
    \subfloat[Confusion matrix for SVC.]{%
        \centering
        \includegraphics[scale=0.5]{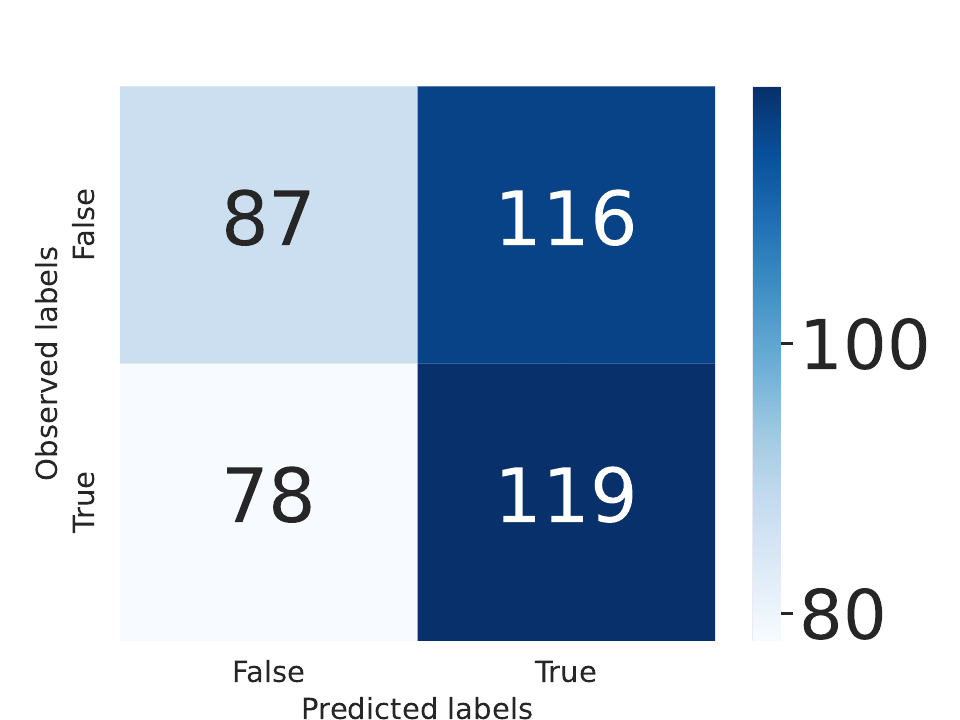}
        \label{fig:cm_mat}
    }
    \hfill
    \subfloat[Metrics for SVC.]{%
        \centering
        \includegraphics[scale=0.25]{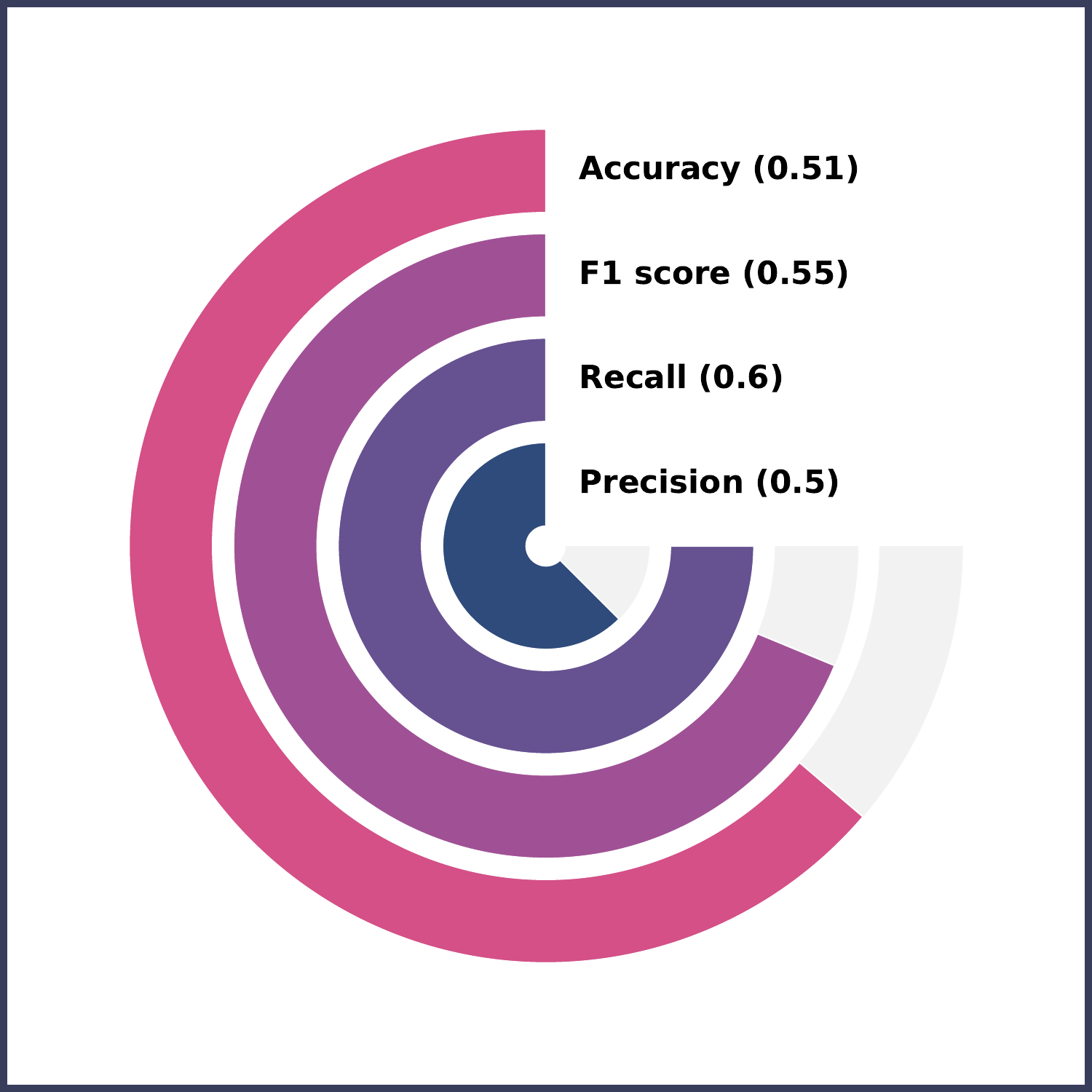}
        \label{fig:radial_bar}
    }
    \caption{Determining if the generated texts through GPT-2 and the test data subset are similar.}
    \label{fig:comparison}
\end{figure*}

As discussed, context $(\textrm{Ctx} = \left[\textrm{ctx}_{1},\textrm{ctx}_{2},\dots,\textrm{ctx}_{n}\right])$ is generated by fine-tuning GPT-2 on the SOCIALIQA dataset's contexts. Once fined-tuned, contexts are generated using just the start and end of text prompts: 
\begin{multline}
    \mathcal{\widehat{\textrm{Ctx}}} = \textrm{fine-tuned GPT2}(\textrm{\texttt{$|<$startoftext$>|$}} \\ \textrm{\texttt{$|<$endoftext$>|$}} )
\end{multline}
where $\mathcal{\widehat{\textrm{Ctx}}}$ is the generated context by fine-tuned GPT-2. For fine-tuning the BART model to generate questions and the possible answer options, BartForConditionalGeneration is used. Figure \ref{fig:model_add_loss} shows the model that has been used in this paper. For every context $\textrm{ctx}_{i}$, the BART model generates text in the following way:
\begin{subequations}\label{eq:prob_formula}
\begin{align}\label{eq:BART_1}
    \textrm{BART}(\textrm{ctx}_{i}+<\textrm{mask}>_{q_{i} \left[:j\right]}) = \nonumber \\\textrm{ctx}_{i} + \mathcal{\widehat{Q}}_{i}
\end{align}
\begin{align}\label{eq:BART_2}
            \textrm{BART}(\textrm{ctx}_{i}+ \mathcal{\widehat{Q}}_{i} + <\textrm{mask}>_{a_{i} \left[:k\right]})  \nonumber \\=\textrm{ctx}_{i} +\mathcal{\widehat{Q}}_{i} + \mathcal{\widehat{A}}_{i}
\end{align}
\begin{align}\label{eq:BART_3}
    \textrm{BART}(\textrm{ctx}_{i}+  \mathcal{Q}_{i}+ \mathcal{\widehat{A}}_{i} + <\textrm{mask}>_{b_{i} \left[:p\right]}) = \nonumber \\\textrm{ctx}_{i} +\mathcal{\widehat{Q}}_{i} + \mathcal{\widehat{A}}_{i}+\mathcal{\widehat{B}}_{i}
\end{align}
\begin{align}\label{eq:BART_4}
    \textrm{BART}(\textrm{ctx}_{i}+  \mathcal{\widehat{Q}}_{i}+ \mathcal{\widehat{A}}_{i} + \mathcal{\widehat{B}}_{i}+<\textrm{mask}>_{c_{i} \left[:r\right]}) = \nonumber \\ \textrm{ctx}_{i} +\mathcal{\widehat{Q}}_{i} + \mathcal{\widehat{A}}_{i}+\mathcal{\widehat{B}}_{i}+\mathcal{\widehat{C}}_{i}
\end{align}
\end{subequations}
where $i$ is the index of the data points, $j$, $k$, $p$ and, $r$ are the total number of tokens for the $i^{th}$ question, option A, option B and, option C, respectively. Further, $\mathcal{\widehat{Q}}_{i}$, $\mathcal{\widehat{A}}_{i}$, $\mathcal{\widehat{B}}_{i}$ and, $\mathcal{\widehat{C}}_{i}$ are the predicted question, option A, option B and option C respectively. As can be seen from Figure \ref{fig:model_add_loss}, the same BART model is fine-tuned with a combination of different training inputs. Equations \ref{eq:BART_1}-\ref{eq:BART_4} give a hierarchical nature to the training where the context is first used to predict the question and so on. For each of these predictions, the cross entropy loss is calculated and added for backpropagation:
\begin{equation}\label{eq:total_loss}
    \mathcal{L} = \mathcal{L}_{q_{i}} + \mathcal{L}_{a_{i}} + \mathcal{L}_{b_{i}} +\mathcal{L}_{c_{i}}
\end{equation}
where $\mathcal{L}_{q_{i}}$, $\mathcal{L}_{a_{i}}$, $\mathcal{L}_{b_{i}}$  and, $\mathcal{L}_{c_{i}}$ are the cross entropy losses calculated from predicting the question, option A, option B, and option C respectively.

This approach was also compared to the GPT-2 generation of the question and three options based on the context. We introduce three special tokens for controlled text generation using GPT-2. In addition to the $|<\textrm{\texttt{startoftext}}>|$ and $|<\textrm{\texttt{endoftext}}>|$ tokens, we also use \texttt{<context>:}, \texttt{<question>:}, \texttt{<ansa>:}, \texttt{<ansb>:} and, \texttt{<ansc>:} to control the generation process of each of the individual parts for each data point. Once these tokens have been added to the tokenizer and the dataset tokenized, we then fine-tune the GPT-2 model and use the cross entropy loss. Each data point before tokenization can be represented as:
\begin{align}
    \mathcal{X}_{i} = \textrm{\texttt{<context>:}} \textrm{ctx}_{i} + \textrm{\texttt{<question>:}} \nonumber \\ \mathcal{Q}_{i} + \textrm{\texttt{<ansa>:}} \mathcal{A}_{i} + \textrm{\texttt{<ansb>:}} \nonumber \\ \mathcal{B}_{i} + \textrm{\texttt{<ansc>:}} \mathcal{C}_{i}  
\end{align}

\section{Results and Discussions}\label{results}

\begin{table*}[h!]
    \centering
    \caption{BERTscore calculated using test data comparing the GPT-2 and BART pipelines for generating the question and three option choices for each context given in the data point.}
    \label{ablation}
    \resizebox{\textwidth}{!}{%
    \begin{tabular}{c|p{3.5cm}|c c c|c c c|c c c}
    \hline
    &\multicolumn{1}{c|}{\textbf{Model used for}} & \multicolumn{3}{c|}{$\mathcal{Q} +\mathcal{A}+\mathcal{B}+\mathcal{C}$} & \multicolumn{3}{c|}{$\mathcal{A}+\mathcal{B}+\mathcal{C}$} & \multicolumn{3}{c|}{$\mathcal{Q}$} \\
    \rule{0pt}{12pt}
    \textbf{Model} & \centering \textbf{BERTscore} & \textbf{Precision} & \textbf{Recall} & \textbf{F-1 score} & \textbf{Precision} & \textbf{Recall} & \textbf{F-1 score} & \textbf{Precision} & \textbf{Recall} & \textbf{F-1 score} \\
    \rule{0pt}{10pt}
     & distilbert-base-uncased & 0.78 & 0.79 & 0.79 & 0.75 & 0.75 & 0.75 & 0.84 & 0.85 & 0.85 \\
    \rule{0pt}{10pt}
    BART & roberta-base & 0.88 & 0.88 & 0.88 & 0.76 & 0.77 & 0.77 & 0.86 & 0.86 & 0.86 \\
    \rule{0pt}{10pt}
     & mnlimicrosoft/deberta-xlarge-mnli & 0.62 & 0.61 & 0.61 & 0.57 & 0.54 & 0.55 & 0.68 & 0.66 & 0.67 \\
    \hline
    \rule{0pt}{10pt}
     & distilbert-base-uncased & 0.71 & 0.76 & 0.73 & 0.70 & 0.74 & 0.72 & 0.80 & 0.83 & 0.81 \\
    \rule{0pt}{10pt}
    GPT-2 & roberta-base & 0.85 & 0.87 & 0.86 & 0.84 & 0.85 & 0.84 & 0.90 & 0.91 & 0.90 \\
    \rule{0pt}{10pt}
     & mnlimicrosoft/deberta-xlarge-mnli & 0.53 & 0.57 & 0.55 & 0.49 & 0.53 & 0.51 & 0.69 & 0.71 & 0.70 \\
    \hline
    \end{tabular}
    }
\end{table*}

As discussed previously, we have used the SOCIALIQA dataset for all of our generation tasks. We used the premium GPU from Google Colab with A100 GPUs for training. The dataset was split into train (75\%), eval (15\%), and test data (10\%). Adam was chosen to be the optimizer. The SOCIALIQA dataset has one context, question, and three
options for every data point. So, we fine-tuned GPT-2 for generating these contexts and then fed the output of the GPT-2 into our BART pipeline explained in the equation \ref{eq:prob_formula} to generate the questions and the options. We compared this pipeline with a similar pipeline, replacing BART with GPT-2 (both pipelines are shown in Figure \ref{fig:model_pipeline}).
 \subsection{Context generation using GPT-2}\label{ctx_gpt}
For generating the context, we fine-tuned a GPT-2 model based on the contexts in the SOCIALIQA dataset. To demonstrate how similar the generated contexts were to the test set, we took one thousand GPT-2 generated texts (after fine-tuning) and the same number of data points from the test set. Further, we used the T-SNE plots to visualize how the test subset and the GPT-2 generated set overlap. We vectorized the generated texts and the subset of the test dataset using scikit-learn's TfidfVectorizer to convert both the test and the generated datasets into a tf-idf feature matrix \cite{tfidf}. Once vectorized, the T-SNE plot was used to visualize the high-dimensional text data into two components. Since we wanted to compare how similar the GPT-2 generated texts and the actual test set, we just used the number of components to differentiate between them. 
As seen from Figure \ref{fig:tsne}, the generated and the test set have a very large overlap. To characterize this overlap, we considered the data points of the scatter plot and used a Support Vector Classifier (SVC) with a radial basis kernel to see if the classifier could separate the data points into two distinct classes. Figure \ref{fig:cm_mat} and \ref{fig:radial_bar} show that the classification accuracy is 0.51, and the F1-score is around 0.55, too. This low F1-score signifies that the classifier could not very distinctly classify the values of the T-SNE plot (essentially the low-dimensional embeddings of the high-dimensional text data) \cite{tfidf}. The low scores on the classifying metrics signify that the text generated by fine-tuning GPT-2 generates data similar to the test set.

\subsection{Evaluation metrics}
\begin{figure*}[h!]
\includegraphics[width=1.02\textwidth]{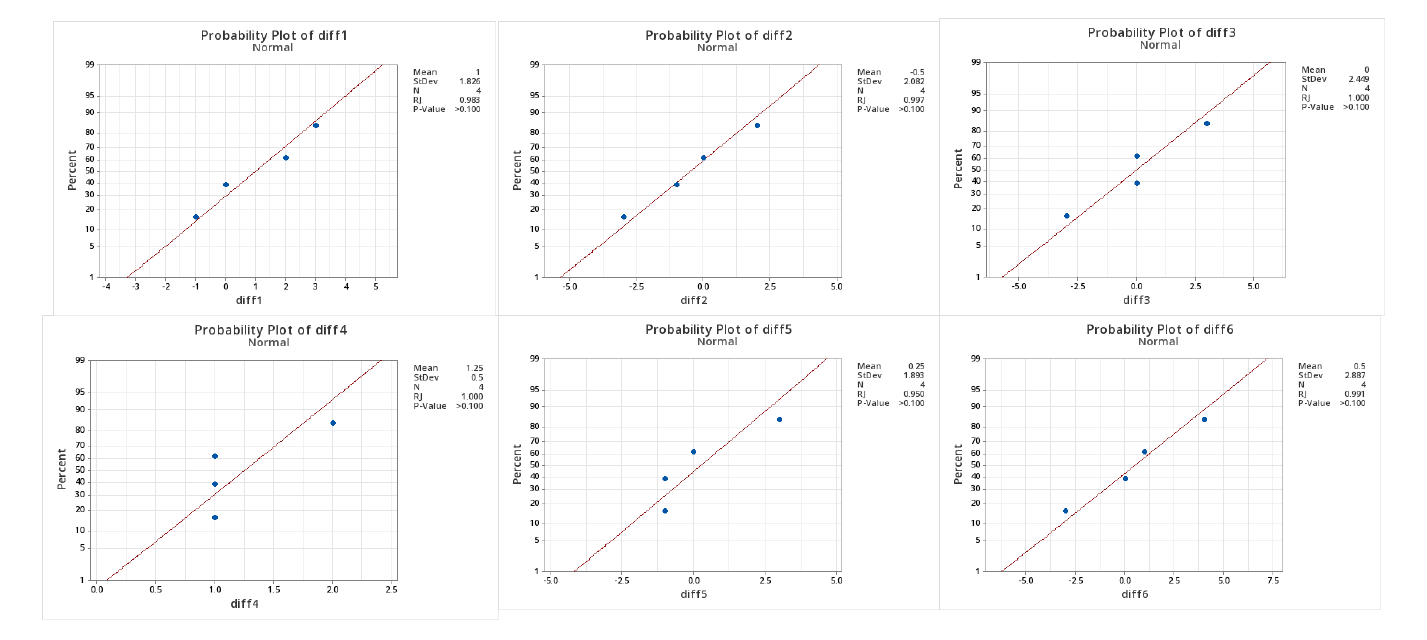}
\caption{Probability plots for Ryan-Joiner normality test based on domain experts' responses on the NASA TLX.} 

\label{fig:nasa_tlx}
\end{figure*}
We use BERTscore to evaluate our question and option generation generation task \cite{zhang2019bertscore}. We do this across generation of questions, options and, both of them combined. There are other metrics used in the literature like the BLEU score \cite{papineni2002bleu}. The reason for using BERTscore is because it uses cosine similarity instead of just making an n-gram based comparison between the actual text and the generated ones. In addition, for calculating the cosine similarities, it uses BERT embeddings between each generated token and every reference token \cite{zhang2019bertscore}. This ensures not just exact phrase matching between the reference and the candidate text but also the contextual relevance between the two. Since there are different BERT based models to calculate the BERTscore from, we choose three of those models: 1) `distilbert-base-uncased', 2) `roberta-base' and, 3) `microsoft/deberta-large-mnlimicrosoft/deberta-xlarge-mnli'. Among these, microsoft/deberta-xlarge-mnli is currently the closest to human evaluations according to \cite{bertscore}. 
We evaluate the BERTscore for generated questions and options for each test sample. As seen from Table \ref{ablation}, our BART pipeline performs better on average for all generation tasks, at least when we use the Microsoft/deberta-xlarge-mnli model for calculating the BERTscore.

\subsection{Self-reports by domain experts}\label{self_reports}
In this section, we conduct statistical analysis on the self-reports collected from the domain experts about the robot session. As discussed previously, their responses were recorded on three surveys: NASA Task Load Index, Godspeed, and a self-curated survey for evaluating the appropriateness of our robotic session for children with ASD.
\subsubsection{NASA TLX}
\begin{figure*}[h!]
    \centering
    \includegraphics[scale = 0.5]{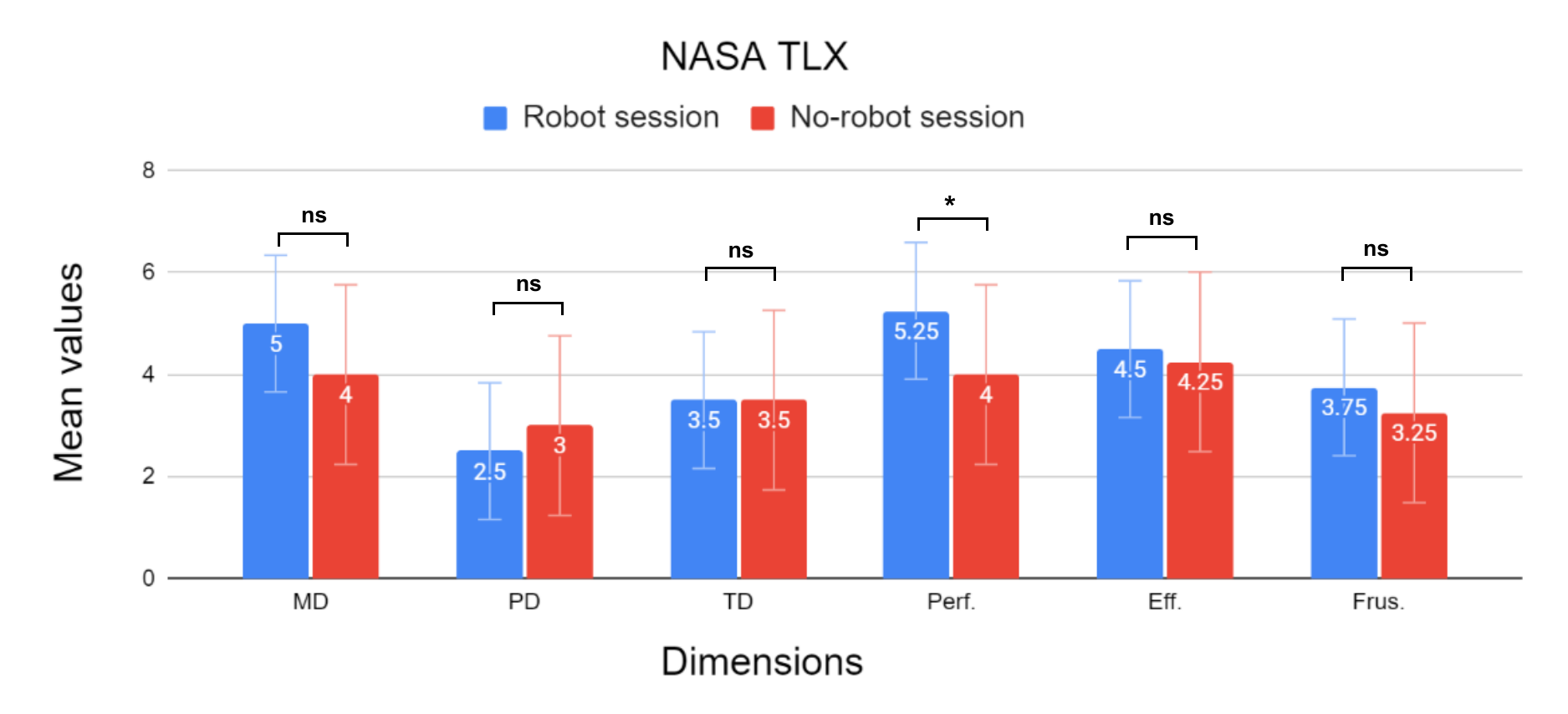}
    \caption{Paired t-test results on the responses of domain experts on the NASA TLX. ns: no significance, $(*)$ denotes $p-$value $<0.05$, and $(**)$ denotes $p-$value $<0.01$.}
    \label{fig:p_val_NASA}
\end{figure*}
We used the NASA TLX to evaluate the domain experts' mental workload (MWL) on six dimensions: mental demand (MD), physical demand (PD), temporal demand (TD), effort (Eff.), performance (Perf.), and frustration (Frus.) levels \cite{hart2006nasa}. The experts were asked to fill the NASA TLX under two different circumstances. In one of the forms, they were asked to fill in the scores on a scale of 1-7 for the robot session. They were asked to do the same on a separate form for a classical perspective-taking session (without the robot). This was done to compare the MWL for the robot vs no-robot sessions.\newline 
\textbf{Test for Normality: } We tested the scores of the four domain experts $\left\{de_{1}, de_{2}, de_{3}, de_{4}\right\}$  for normality using Ryan-Joiner (similar to Shapiro-Wilk test) test for normality in Minitab. The difference in the ratings of $de_{N,h}$, where $h \in \left\{1,2,3,4\right\}$ for the robot vs. no robot sessions were used to check for normality as given in equation \ref{eq:norm_check_rep}:
\begin{align}\label{eq:norm_check_rep}
    \textrm{diff}_{h} = de_{N,h,r} - de_{N,h,nr} \nonumber \\ 
    \mathcal{D}_{N,e} = \left\{\textrm{diff}_{1,e},\textrm{diff}_{2,e}, \textrm{diff}_{3,e}, \textrm{diff}_{4,e}\right\} \nonumber \\
    \textrm{Check if: }  \mathcal{D}_{N,e} \sim \mathcal{N}(\mu_{N,e},\sigma_{N,e})
\end{align}
where $de_{N,h,r}$ is the $h^{th}$ domain expert's rating on the NASA TLX for the robot session whereas  $de_{N,h,nr}$ represents the same for the no-robot session, and $e \in \left\{\textrm{MD}, \textrm{PD}, \textrm{TD}, \textrm{Perf.}, \textrm{Eff.}, \textrm{Frus.}\right\}$. 
The hypothesis for the Ryan-Joiner test is:
\begin{align}\label{eq:test_norm}
    \textrm{Null hypothesis (H}_{0}): \textrm{data follows} \nonumber \\ \nonumber \textrm{normal distribution}
    \nonumber \\
    \textrm{Alternate hypothesis (H}_{a}): \textrm{data does not follow} \nonumber \\ \textrm{normal distribution}
\end{align}
Since the p-value $>0.100$, for each of these dimensions (MD - Perf.), we accept the Null Hypothesis ($H_{0}$) that the distribution is normal.
\newline
\textbf{Paired t-test: }Since $\mathcal{D}_{N}$ is normally distributed, it fulfills the assumption of the paired t-test. Hence, we compare the responses of the experts between the robot and no-robot sessions based on the following hypothesis:

\begin{align}\label{eq:nasa_t_test}
    \textrm{Null hypothesis (H}_{0}): \mu_{r} = \mu_{nr} \nonumber \\
 \textrm{Alternate hypothesis (H}_{a}): \mu_{r} \neq \mu_{nr} 
\end{align}
where $\mu_{r}$ denotes the mean of the robot session and $\mu_{nr}$ denotes the mean of the no-robot session.

As seen from Figure \ref{fig:p_val_NASA}, there is no significant increase in the MWL of the domain experts when they are operating the robot. This is because the $p-$value $> 0.05$ for MS, PD, TD, Eff., and Frus., rejecting $H_{a}$ in equation \ref{eq:nasa_t_test}. In addition to this observation, we can also see from Figure \ref{fig:p_val_NASA} that the $p-$value $<0.05$ for Perf. and hence we reject $H_{0}$ from equation \ref{eq:nasa_t_test}.
However, as can be seen from Figure \ref{fig:p_val_NASA}, the performance in the case of the robot session is higher than the no-robot session. The mean performance value was recorded as $5.250$, compared to the $4.00$ in the no-robot session. Based on this comparison of means, we can conclude that the robot session does not demand any statistically significant additional MWL while being a scalable solution due to using LLMs. Moreover, the performance dimension of the NASA TLX shows a statistically significant increment in performance during the robot session compared to the no-robot session, which makes a strong case for domain experts to adopt such a system. 
\begin{figure*}
    \centering
    \includegraphics[width=1.06\textwidth]{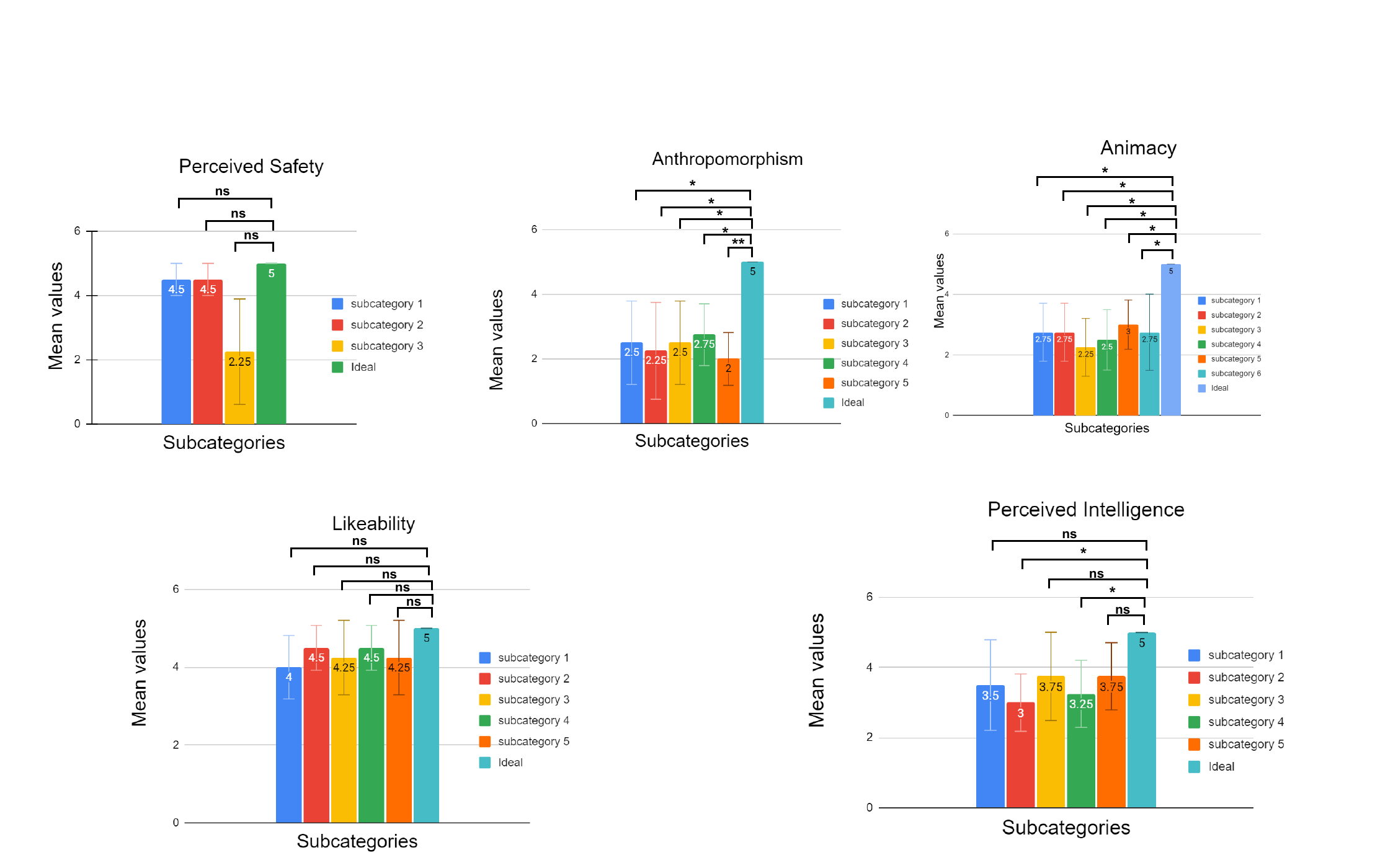}
    \caption{Paired t-test results of the perceived features of the domain experts on the Godspeed survey. ns: no significance, $(*)$ denotes $p-$value $<0.05$, and $(**)$ denotes $p-$value $<0.01$.}
    \label{fig:p_val_godspeed}
\end{figure*}
\subsubsection{Godspeed}
To evaluate the perception of the domain experts about the robot, we asked them to fill in the Godspeed survey \cite{bartneck2009measurement}. These perceptions are evaluated across five perceived features: 
\begin{itemize}
    \item Perceived safety (PS)
    \begin{itemize}
        \item anxious to relaxed
        \item agitated to calm
        \item quiescent to surprised
    \end{itemize}
    \item Anthropomorphism (AP)
    \begin{itemize}
        \item fake to natural
        \item machinelike to humanlike
        \item unconscious to conscious
        \item artificial to lifelike
        \item moving rigidly to moving elegantly
    \end{itemize}
    \item Animacy (AN)
    \begin{itemize}
        \item dead to alive
        \item stagnant to lively
        \item mechanical to organic 
        \item artificial to lifelike 
        \item inert to interactive 
        \item apathetic to responsive 
    \end{itemize}
    \item Likeability (LK)
\begin{itemize}
    \item dislike to like 
    \item unfriendly to friendly 
    \item unkind to kind 
    \item unpleasant to pleasant
    \item awful to nice
\end{itemize}
\item Perceived Intelligence (PI)
\begin{itemize}
    \item incompetent to competant 
    \item ignorant to knowledgeable
    \item irresponsible to responsible 
    \item unintelligent to intelligent
    \item foolish to sensible
\end{itemize}
\end{itemize}

Each of these perceived features,  $ \mathcal{G}= \left\{\textrm{PS}, \textrm{AP}, \textrm{AN}, \textrm{LK}, \textrm{PI} \right\}$  has different sub-categories. For example, perceived safety has three sub-categories. Each of these subcategories is rated by the domain experts on a scale of 1-5. 
\newline
\begin{figure*}
    \centering
    \includegraphics[scale=0.5]{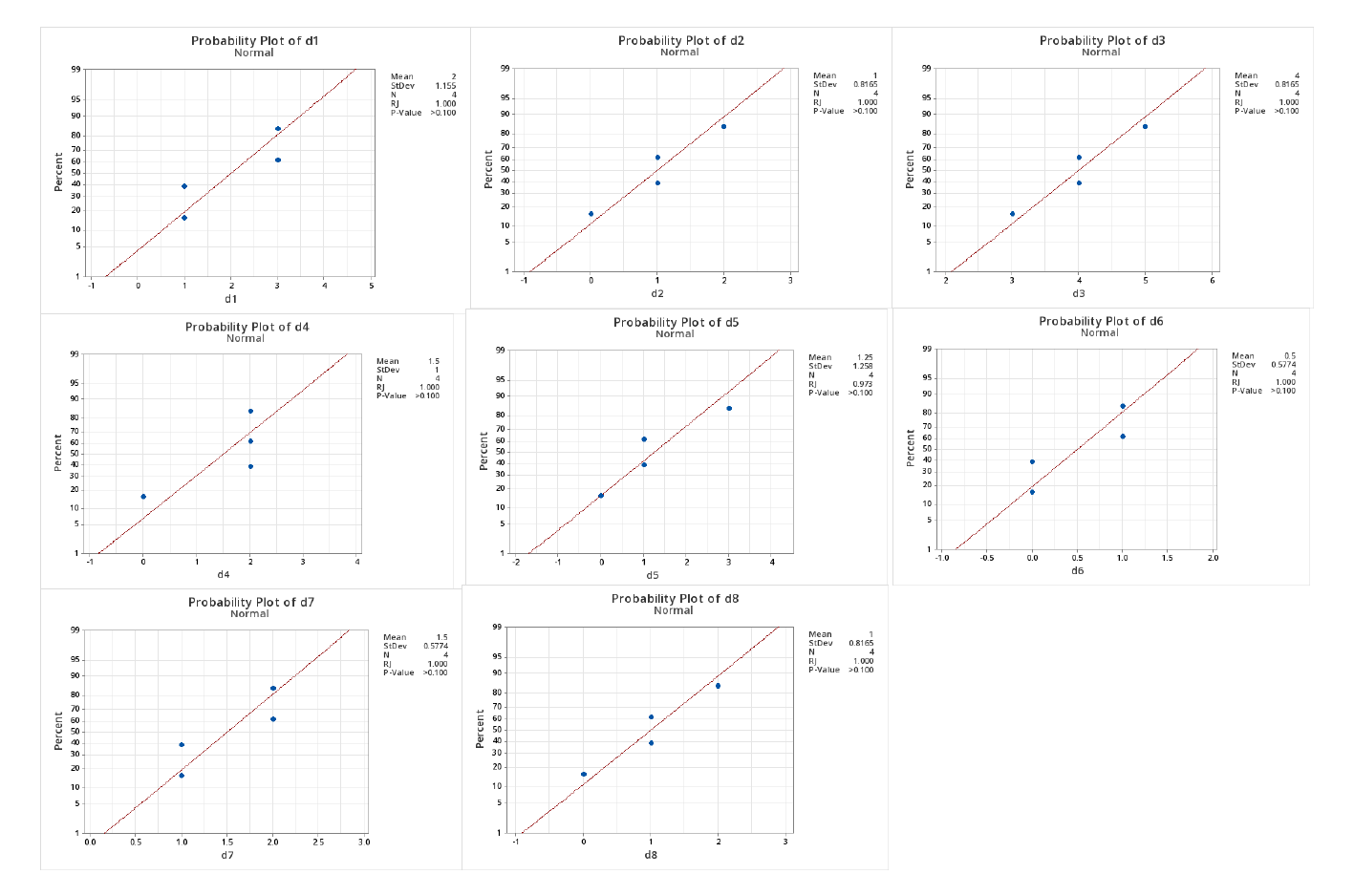}
    \caption{Normality test of the domain experts' scores on the survey evaluating our human-mediated autonomous system for children with ASD.}
    \label{fig:app_norm}
\end{figure*}
\textbf{Test for normality: }
We follow the same hypothesis as in equation \ref{eq:test_norm}. The scores of the domain experts is represented by $de_{h,s}^{g}$, where $g \in \mathcal{G}$, $s \in \mathcal{S}^{g}$, and $\mathcal{S}^{g}$ represents the set of all subcategories of perceived feature $g \in \mathcal{G}$, and the ideal scores are represented by $\mathcal{I} = \left\{5,5,5,5\right\}$. 
\begin{align}
    \textrm{diff}_{g,h,s} = de_{h,s}^{g} - \mathcal{I} \nonumber\\
    \mathcal{D}_{g,s} = \left\{\textrm{diff}_{g,1,s},\textrm{diff}_{g,2,s},\textrm{diff}_{g,3,s},\textrm{diff}_{g,4,s}\right\} \nonumber \\
    \textrm{Check: } \mathcal{D}_{g,s} \sim \mathcal{N}(\mu_{g,s},\sigma_{g,s})
\end{align}
Using the same Ryan-Joiner test for normality, and the hypothesis as shown in equation \ref{eq:test_norm}, we find $p-$value $>0.100$ for $\mathcal{D}_{g,s}$, hence rejecting $H_{a}$ (distribution is normal).
\newline
\textbf{Paired t-test: }
\begin{align}\label{eq:godspeed_t_test}
    \textrm{Null hypothesis (H}_{0}): \mu_{s,Go}^{g} = \mu_{i,Go} \nonumber \\
 \textrm{Alternate hypothesis (H}_{1}): \mu_{s,Go}^{g} \neq \mu_{i,Go} 
\end{align}
where, $ \mu_{s,Go}^{g}$ is the mean value of $s \in S^{g}$ of the perceived feature $g \in \mathcal{G}$, and $\mu_{i,Go}$ is the mean value of the ideal case i.e. $\mathcal{I}$. Figure \ref{fig:p_val_godspeed} shows the results of the paired t-test. On comparing the mean values of the domain experts' scores with the ideal case scenario, we reject $H_{0}$ for all subcategories for anthropomorphism and animacy and two subcategories for perceived intelligence. However, there was no significant statistical difference ($p-$value $>0.05$) in the mean scores for every subcategory of perceived safety and likeability with the ideal case scenario. This means the robot was ideally perceived to be safe and likeable by the domain experts on the Godspeed surveys. 

\subsection{Appropriateness for children with ASD}
This survey was made specifically to take the domain experts' opinions to evaluate the appropriateness of our human mediated robotic intervention. We ask them eight questions to include the aspects of our robotic session not covered in the previous two surveys:
\begin{itemize}
    \item \textbf{Q1: } How would you rate the roles of the robot as a initiator (content generator)? 
    \item \textbf{Q2: } How would you rate the roles of the robot as a prompter?
    \item \textbf{Q3: } How would you rate the role of the robot as a reinforcer?
    \item \textbf{Q4: } How appropriate was this entire session with the robot acting as a provoker, prompter, and reinforcer for children with ASD?
    \item \textbf{Q5: } Would you prefer this human-mediated autonomous system as compared to a completely teleoperated system?
    \item \textbf{Q6: } Did you find the robot reliable (with no technical glitches) through the entire duration of the session?
    \item \textbf{Q7: } Would you prefer this human-mediated autonomous system in a therapeutic or educational settings?
    \item \textbf{Q8: } Did you find the Graphical Use Interface (GUI) helpful to operate the robot?
\end{itemize}
For each question, Q$_{i}$ from $\mathcal{Q} = \left\{Q_{1}, Q_{2}, \dots, Q_{8}\right\}$, we ask the domain experts to give their ratings on a scale of 1-5. \newline
\textbf{Test for normality: } For a question $q \in \mathcal{Q}$, the scores of the domain experts $de_{h,q}$ were compared against the ideal scores $\mathcal{I} = \left\{5,5,5,5\right\}$.
\begin{align}
    \textrm{d}_{h,q} = de_{h,q} - \mathcal{I} \nonumber \\
    D_{q} =\left\{\textrm{d}_{1,q},\textrm{d}_{2,q},\textrm{d}_{3,q},\textrm{d}_{4,q}\right\} \nonumber \\
    \textrm{Check if: } D_{q} \sim \mathcal{N}(\mu_{q}, \sigma_{q})
\end{align}
The hypothesis used in equation \ref{eq:test_norm} were used to conduct the Ryan-Joiner test for normality. It was found that for all $D_{q}$, the $p-$value $>0.100$, hence accepting H$_{0}$ that the distribution is normal.
\begin{figure}[h!]
    \centering
    \includegraphics[scale = 0.455]{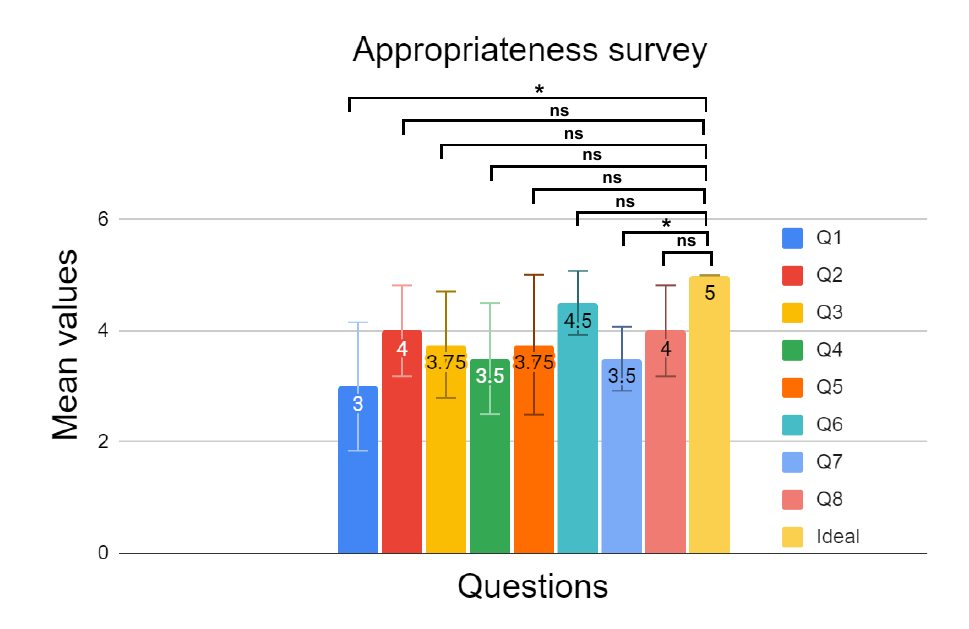}
    \caption{Paired t-test results for the appropriateness survey.}
    \label{fig:appt_sur}
\end{figure}
\newline
\textbf{Paired t-test: } 
\begin{align}\label{eq:appt_sur_hypo}
    \textrm{Null hypothesis (H}_{0}): \mu_{q,A} = \mu_{i,A} \nonumber \\
 \textrm{Alternate hypothesis (H}_{1}): \mu_{q,A} \neq \mu_{i,A} 
\end{align}
where $\mu_{q,A}$ is the mean of the domain experts' responses on $q \in \mathcal{Q}$ and $\mu_{i,A}$ is the mean for the ideal case scenario $\mathcal{I}$. From Figure \ref{fig:appt_sur}, it can be seen that the roles of the robot for the entire session acting as an initiator, prompter, and reinforcer (Q4) were statistically no different from the ideal case scenario ($p-$value $>0.05$). In addition, the roles of the robot as a prompter and reinforcer individually too were not statistically significant from the ideal case scenario ($p-$value $>0.05$). Further, based on the same rational of comparing the means of the domain experts scores to the ideal case scenario, the domain experts preferred our human-mediated autonomous system compared to a completely teleoperated system (Q5). This is because our human mediated system was reliable during the time the robot was being operated by the domain expert. It faced practically no technical glitches that would hinder the progress of the session (Q6). This can be seen from ($p-$value $>0.05$) for Q6. To make the system more user friendly, the GUI as shown in Figure \ref{fig:gui} was also considered very helpful to operate the robot (Q8 and Figure \ref{fig:appt_sur}).    
\begin{figure}[h!]
    \centering
    \includegraphics[scale = 0.45]{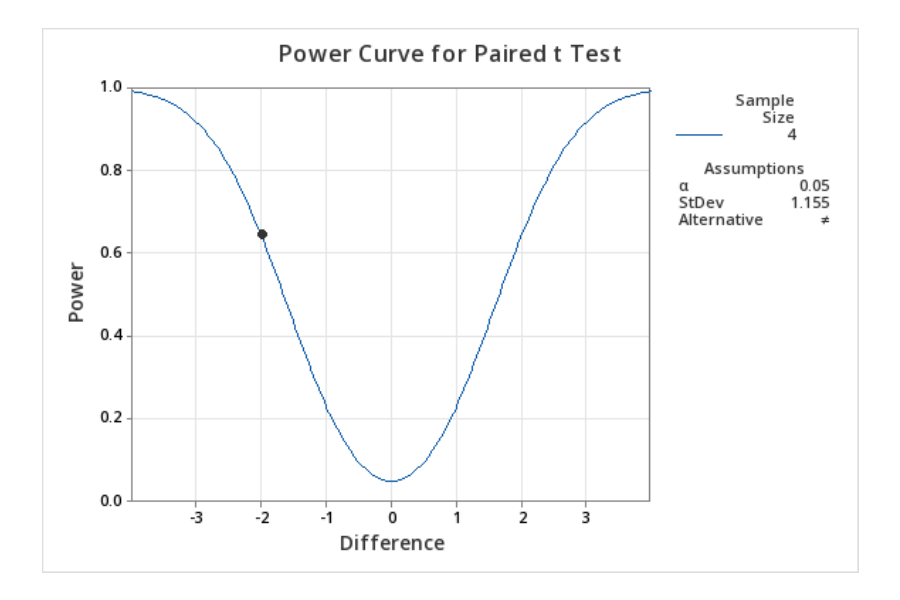}
    \caption{Power of paired t-test for Q1.}
    \label{fig:power}
\end{figure}
\begin{figure}
    \centering
    \includegraphics[scale = 0.45]{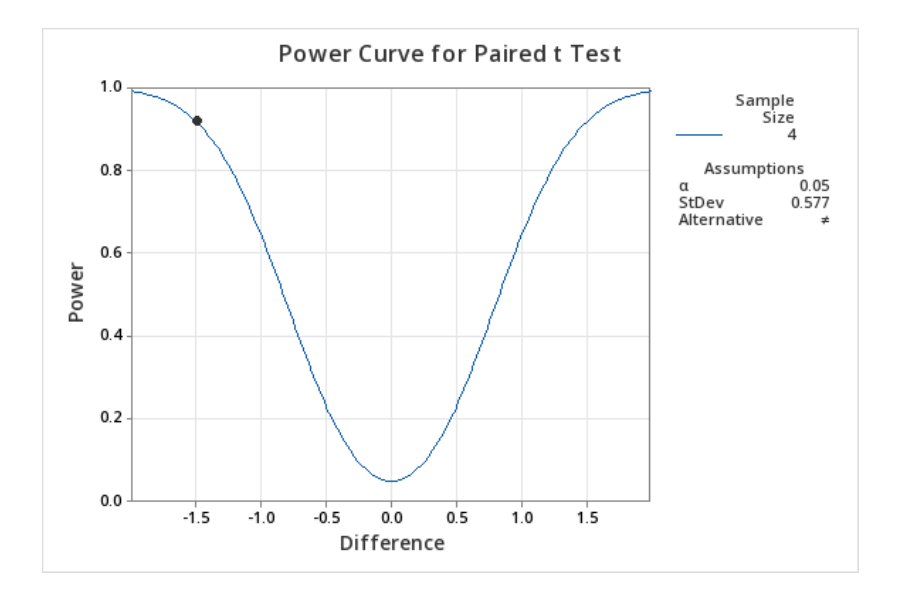}
    \caption{Power of paired t-test for Q7}
    \label{fig:enter-label}
\end{figure}
The scores for Q1 and Q7 came out to be statistically different from the ideal case scenario. However, for Q1, an important point to note is that the power of the paired t-test is 0.6433 (see Figure \ref{fig:power}). So, because of this low power, it is difficult to reliably reject the null hypothesis in equation \ref{eq:appt_sur_hypo}. 
But for Q7, the power comes out to be 0.917. Although for Q7, the mean is statistically different from the ideal case scenario, the mean for Q7 is 4.00 which is more than half the length of the 5-point likert scale. In addition, the low score of Q7 can be attributed to the use of a dataset (SOCIALIQA) (used fine-tuning our LLM approach) that was not built keeping in mind children with ASD. However, this human mediated autonomous system mitigates this effect since the authority to override the robot lies with the domain expert (explained in Section \ref{human_mediated_aut}).

\section{Limitations and Future work}\label{limit_future}
Although we could make conclusions based on the domain experts' responses on different surveys, it will be more helpful to collect data from more such experts. This would probably help us get different results, such as in Figure \ref{fig:appt_sur}. In Figure \ref{fig:appt_sur}, except for Q1 and Q7, all other responses were statistically similar to the ideal cases. Furthermore, having more number of domain experts to rate might impact the results of the paired t-test shown in Figure \ref{fig:p_val_godspeed}.

Beyond the self reports, we would like to record naturalistic conversations between domain experts and children with ASD. This will help us achieve multiple goals. First, we can extend our current work by fine-tuning LLMs into making natural conversations with the child with ASD. Second, we can create a dataset of the communications recorded between the domain experts and the robot which would help us tailor our LLM models to specific scenarios, thus personalizing this robot experience for children with ASD. 

\section{Conclusion}\label{conclusion}
In this paper, we develop an intervention scheme for children with ASD and introduce autonomy in one of the elements of the pipeline, i.e. content generation. This helps make the approach more scalable in the sense that an LLM pipeline can be fine-tuned to generate contents specific to a situation, or to a therapeutic goal or outcome in mind. The NAO robot used in this paper acts as a initiator, stimulator, and a reinforcer during the intervention. This LLM pipeline contributes to the autonomy in the content generation for the initiator and the prompter roles of the robot. The decision on which role the robot should take up during the intervention lies with the human, hence contributing the human-mediated autonomy of the system. This gives control to the domain expert on what contents generated by the LLM pipeline is appropriate for the robot to speak during an intervention of children with ASD. 

In addition to developing a technical pipline for a human-mediated autonomous system, we also conducted a user study to collect the opinion of domain experts. We do that by conducting a ten minute session with the robot, domain expert, and an actor participant. We collect the opinions on the domain experts based on three surveys: NASA TLX, Godspeed, and an appropriateness survey. We perform statistical tests to compare the responses of the domain experts and report it in Section \ref{self_reports}.

\section{Ethics approval}
This study was approved by the University's Institute Review Board (IRB). All the participants gave informed consent, and they had an option to opt out of the study at any stage.


\bibliographystyle{IEEEtran}
\bibliography{references.bib}

\begin{thebibliography}{10}
\providecommand{\url}[1]{#1}
\csname url@samestyle\endcsname
\providecommand{\newblock}{\relax}
\providecommand{\bibinfo}[2]{#2}
\providecommand{\BIBentrySTDinterwordspacing}{\spaceskip=0pt\relax}
\providecommand{\BIBentryALTinterwordstretchfactor}{4}
\providecommand{\BIBentryALTinterwordspacing}{\spaceskip=\fontdimen2\font plus
\BIBentryALTinterwordstretchfactor\fontdimen3\font minus \fontdimen4\font\relax}
\providecommand{\BIBforeignlanguage}[2]{{%
\expandafter\ifx\csname l@#1\endcsname\relax
\typeout{** WARNING: IEEEtran.bst: No hyphenation pattern has been}%
\typeout{** loaded for the language `#1'. Using the pattern for}%
\typeout{** the default language instead.}%
\else
\language=\csname l@#1\endcsname
\fi
#2}}
\providecommand{\BIBdecl}{\relax}
\BIBdecl

\bibitem{lambert2020systematic}
A.~Lambert, N.~Norouzi, G.~Bruder, and G.~Welch, ``A systematic review of ten years of research on human interaction with social robots,'' \emph{International Journal of Human--Computer Interaction}, vol.~36, no.~19, pp. 1804--1817, 2020.

\bibitem{forlizzi2006service}
J.~Forlizzi and C.~DiSalvo, ``Service robots in the domestic environment: a study of the roomba vacuum in the home,'' in \emph{Proceedings of the 1st ACM SIGCHI/SIGART conference on Human-robot interaction}, 2006, pp. 258--265.

\bibitem{kidd2008robots}
C.~D. Kidd and C.~Breazeal, ``Robots at home: Understanding long-term human-robot interaction,'' in \emph{2008 IEEE/RSJ International Conference on Intelligent Robots and Systems}.\hskip 1em plus 0.5em minus 0.4em\relax IEEE, 2008, pp. 3230--3235.

\bibitem{tsui2011exploring}
K.~M. Tsui, M.~Desai, H.~A. Yanco, and C.~Uhlik, ``Exploring use cases for telepresence robots,'' in \emph{Proceedings of the 6th international conference on Human-robot interaction}, 2011, pp. 11--18.

\bibitem{okamura2004methods}
A.~M. Okamura, ``Methods for haptic feedback in teleoperated robot-assisted surgery,'' \emph{Industrial Robot: An International Journal}, vol.~31, no.~6, pp. 499--508, 2004.

\bibitem{el2020review}
I.~El~Rassi and J.-M. El~Rassi, ``A review of haptic feedback in tele-operated robotic surgery,'' \emph{Journal of medical engineering \& technology}, vol.~44, no.~5, pp. 247--254, 2020.

\bibitem{bartl2021robot}
K.~D. Bartl-Pokorny, M.~Pyka{\l}a, P.~Uluer, D.~E. Barkana, A.~Baird, H.~Kose, T.~Zorcec, B.~Robins, B.~W. Schuller, and A.~Landowska, ``Robot-based intervention for children with autism spectrum disorder: a systematic literature review,'' \emph{IEEE Access}, vol.~9, pp. 165\,433--165\,450, 2021.

\bibitem{saral2022autism}
D.~Saral, S.~Olcay, and H.~Ozturk, ``Autism spectrum disorder: When there is no cure, there are countless of treatments,'' \emph{Journal of Autism and Developmental Disorders}, pp. 1--16, 2022.

\bibitem{regier2013dsm}
D.~A. Regier, E.~A. Kuhl, and D.~J. Kupfer, ``The dsm-5: Classification and criteria changes,'' \emph{World psychiatry}, vol.~12, no.~2, pp. 92--98, 2013.

\bibitem{saral2023autism}
D.~Saral, S.~Olcay, and H.~Ozturk, ``Autism spectrum disorder: When there is no cure, there are countless of treatments,'' \emph{Journal of Autism and Developmental Disorders}, vol.~53, no.~12, pp. 4901--4916, 2023.

\bibitem{steinbrenner2020evidence}
J.~R. Steinbrenner, K.~Hume, S.~L. Odom, K.~L. Morin, S.~W. Nowell, B.~Tomaszewski, S.~Szendrey, N.~S. McIntyre, S.~Y{\"u}cesoy-{\"O}zkan, and M.~N. Savage, ``Evidence-based practices for children, youth, and young adults with autism.'' \emph{FPG child development institute}, 2020.

\bibitem{thabtah2020new}
F.~Thabtah and D.~Peebles, ``A new machine learning model based on induction of rules for autism detection,'' \emph{Health informatics journal}, vol.~26, no.~1, pp. 264--286, 2020.

\bibitem{ntaountaki2019robotics}
P.~Ntaountaki, G.~Lorentzou, A.~Lykothanasi, P.~Anagnostopoulou, V.~Alexandropoulou, and A.~Drigas, ``Robotics in autism intervention.'' \emph{Int. J. Recent Contributions Eng. Sci. IT}, vol.~7, no.~4, pp. 4--17, 2019.

\bibitem{wijayasinghe2016human}
I.~B. Wijayasinghe, I.~Ranatunga, N.~Balakrishnan, N.~Bugnariu, and D.~O. Popa, ``Human--robot gesture analysis for objective assessment of autism spectrum disorder,'' \emph{International Journal of Social Robotics}, vol.~8, pp. 695--707, 2016.

\bibitem{scassellati2018improving}
B.~Scassellati, L.~Boccanfuso, C.-M. Huang, M.~Mademtzi, M.~Qin, N.~Salomons, P.~Ventola, and F.~Shic, ``Improving social skills in children with asd using a long-term, in-home social robot,'' \emph{Science Robotics}, vol.~3, no.~21, p. eaat7544, 2018.

\bibitem{begum2015measuring}
M.~Begum, R.~W. Serna, D.~Kontak, J.~Allspaw, J.~Kuczynski, H.~A. Yanco, and J.~Suarez, ``Measuring the efficacy of robots in autism therapy: How informative are standard hri metrics','' in \emph{Proceedings of the Tenth Annual ACM/IEEE International Conference on Human-Robot Interaction}, 2015, pp. 335--342.

\bibitem{cao2019robot}
H.-L. Cao, P.~G. Esteban, M.~Bartlett, P.~Baxter, T.~Belpaeme, E.~Billing, H.~Cai, M.~Coeckelbergh, C.~Costescu, D.~David \emph{et~al.}, ``Robot-enhanced therapy: Development and validation of supervised autonomous robotic system for autism spectrum disorders therapy,'' \emph{IEEE robotics \& automation magazine}, vol.~26, no.~2, pp. 49--58, 2019.

\bibitem{van2020adherence}
I.~van~den Berk-Smeekens, M.~van Dongen-Boomsma, M.~W. De~Korte, J.~C. Den~Boer, I.~J. Oosterling, N.~C. Peters-Scheffer, J.~K. Buitelaar, E.~I. Barakova, T.~Lourens, W.~G. Staal \emph{et~al.}, ``Adherence and acceptability of a robot-assisted pivotal response treatment protocol for children with autism spectrum disorder,'' \emph{Scientific reports}, vol.~10, no.~1, p. 8110, 2020.

\bibitem{takata2023social}
K.~Takata, Y.~Yoshikawa, T.~Muramatsu, Y.~Matsumoto, H.~Ishiguro, M.~Mimura, and H.~Kumazaki, ``Social skills training using multiple humanoid robots for individuals with autism spectrum conditions,'' \emph{Frontiers in Psychiatry}, vol.~14, 2023.

\bibitem{boccanfuso2017low}
L.~Boccanfuso, S.~Scarborough, R.~K. Abramson, A.~V. Hall, H.~H. Wright, and J.~M. O’Kane, ``A low-cost socially assistive robot and robot-assisted intervention for children with autism spectrum disorder: field trials and lessons learned,'' \emph{Autonomous Robots}, vol.~41, pp. 637--655, 2017.

\bibitem{pennisi2016autism}
P.~Pennisi, A.~Tonacci, G.~Tartarisco, L.~Billeci, L.~Ruta, S.~Gangemi, and G.~Pioggia, ``Autism and social robotics: A systematic review,'' \emph{Autism Research}, vol.~9, no.~2, pp. 165--183, 2016.

\bibitem{stolarz2022personalized}
M.~Stolarz, A.~Mitrevski, M.~Wasil, and P.~G. Pl{\"o}ger, ``Personalized behaviour models: A survey focusing on autism therapy applications,'' in \emph{Lifelong Learning and Personalization in Long-Term Human-Robot Interaction (LEAP-HRI), March 7, 2022, Virtual, as part of the 16th ACM/IEEE International Conference on Human-Robot Interaction (HRI 2022)}, 2022.

\bibitem{cabibihan2013robots}
J.-J. Cabibihan, H.~Javed, M.~Ang, and S.~M. Aljunied, ``Why robots? a survey on the roles and benefits of social robots in the therapy of children with autism,'' \emph{International journal of social robotics}, vol.~5, pp. 593--618, 2013.

\bibitem{salimi2021social}
Z.~Salimi, E.~Jenabi, and S.~Bashirian, ``Are social robots ready yet to be used in care and therapy of autism spectrum disorder: A systematic review of randomized controlled trials,'' \emph{Neuroscience \& Biobehavioral Reviews}, vol. 129, pp. 1--16, 2021.

\bibitem{holeva2022effectiveness}
V.~Holeva, V.~Nikopoulou, C.~Lytridis, C.~Bazinas, P.~Kechayas, G.~Sidiropoulos, M.~Papadopoulou, M.~Kerasidou, C.~Karatsioras, N.~Geronikola \emph{et~al.}, ``Effectiveness of a robot-assisted psychological intervention for children with autism spectrum disorder,'' \emph{Journal of Autism and Developmental Disorders}, pp. 1--17, 2022.

\bibitem{huijnen2019roles}
C.~A. Huijnen, M.~A. Lexis, R.~Jansens, and L.~P. de~Witte, ``Roles, strengths and challenges of using robots in interventions for children with autism spectrum disorder (asd),'' \emph{Journal of autism and developmental disorders}, vol.~49, pp. 11--21, 2019.

\bibitem{scassellati2012robots}
B.~Scassellati, H.~Admoni, and M.~Matari{\'c}, ``Robots for use in autism research,'' \emph{Annual review of biomedical engineering}, vol.~14, pp. 275--294, 2012.

\bibitem{baron1985does}
S.~Baron-Cohen, A.~M. Leslie, and U.~Frith, ``Does the autistic child have a “theory of mind”?'' \emph{Cognition}, vol.~21, no.~1, pp. 37--46, 1985.

\bibitem{gould2011teaching}
E.~Gould, J.~Tarbox, D.~O'Hora, S.~Noone, and R.~Bergstrom, ``Teaching children with autism a basic component skill of perspective-taking,'' \emph{Behavioral Interventions}, vol.~26, no.~1, pp. 50--66, 2011.

\bibitem{welsh2019teaching}
F.~Welsh, A.~C. Najdowski, D.~Strauss, L.~Gallegos, and J.~A. Fullen, ``Teaching a perspective-taking component skill to children with autism in the natural environment,'' \emph{Journal of Applied Behavior Analysis}, vol.~52, no.~2, pp. 439--450, 2019.

\bibitem{kimhi2014theory}
Y.~Kimhi, ``Theory of mind abilities and deficits in autism spectrum disorders,'' \emph{Topics in Language Disorders}, vol.~34, no.~4, pp. 329--343, 2014.

\bibitem{hoddenbach2012individual}
E.~Hoddenbach, H.~M. Koot, P.~Clifford, C.~Gevers, C.~Clauser, F.~Boer, and S.~Begeer, ``Individual differences in the efficacy of a short theory of mind intervention for children with autism spectrum disorder: a randomized controlled trial,'' \emph{Trials}, vol.~13, no.~1, pp. 1--7, 2012.

\bibitem{begeer2015effects}
S.~Begeer, P.~Howlin, E.~Hoddenbach, C.~Clauser, R.~Lindauer, P.~Clifford, C.~Gevers, F.~Boer, and H.~M. Koot, ``Effects and moderators of a short theory of mind intervention for children with autism spectrum disorder: A randomized controlled trial,'' \emph{Autism Research}, vol.~8, no.~6, pp. 738--748, 2015.

\bibitem{rudovic2018personalized}
O.~Rudovic, J.~Lee, M.~Dai, B.~Schuller, and R.~W. Picard, ``Personalized machine learning for robot perception of affect and engagement in autism therapy,'' \emph{Science Robotics}, vol.~3, no.~19, p. eaao6760, 2018.

\bibitem{vaswani2017attention}
A.~Vaswani, N.~Shazeer, N.~Parmar, J.~Uszkoreit, L.~Jones, A.~N. Gomez, {\L}.~Kaiser, and I.~Polosukhin, ``Attention is all you need,'' \emph{Advances in neural information processing systems}, vol.~30, 2017.

\bibitem{ezen2020comparison}
A.~Ezen-Can, ``A comparison of lstm and bert for small corpus,'' \emph{arXiv e-prints}, pp. arXiv--2009, 2020.

\bibitem{rahali2023deeppress}
A.~Rahali and M.~A. Akhloufi, ``Deeppress: guided press release topic-aware text generation using ensemble transformers,'' \emph{Neural Computing and Applications}, vol.~35, no.~17, pp. 12\,847--12\,874, 2023.

\bibitem{le2016quantifying}
P.~Le and W.~Zuidema, ``Quantifying the vanishing gradient and long distance dependency problem in recursive neural networks and recursive lstms,'' in \emph{Proceedings of the 1st Workshop on Representation Learning for NLP}, 2016, pp. 87--93.

\bibitem{hochreiter1998vanishing}
S.~Hochreiter, ``The vanishing gradient problem during learning recurrent neural nets and problem solutions,'' \emph{International Journal of Uncertainty, Fuzziness and Knowledge-Based Systems}, vol.~6, no.~02, pp. 107--116, 1998.

\bibitem{bahdanau2014neural}
D.~Bahdanau, K.~Cho, and Y.~Bengio, ``Neural machine translation by jointly learning to align and translate,'' \emph{arXiv preprint arXiv:1409.0473}, 2014.

\bibitem{devlin2018bert}
J.~Devlin, M.-W. Chang, K.~Lee, and K.~Toutanova, ``Bert: Pre-training of deep bidirectional transformers for language understanding,'' \emph{arXiv preprint arXiv:1810.04805}, 2018.

\bibitem{chan-fan-2019-bert}
\BIBentryALTinterwordspacing
Y.-H. Chan and Y.-C. Fan, ``{BERT} for question generation,'' in \emph{Proceedings of the 12th International Conference on Natural Language Generation}.\hskip 1em plus 0.5em minus 0.4em\relax Tokyo, Japan: Association for Computational Linguistics, Oct.{--}Nov. 2019, pp. 173--177. [Online]. Available: \url{https://aclanthology.org/W19-8624}
\BIBentrySTDinterwordspacing

\bibitem{chan2019recurrent}
------, ``A recurrent bert-based model for question generation,'' in \emph{Proceedings of the 2nd workshop on machine reading for question answering}, 2019, pp. 154--162.

\bibitem{luo2022biogpt}
R.~Luo, L.~Sun, Y.~Xia, T.~Qin, S.~Zhang, H.~Poon, and T.-Y. Liu, ``Biogpt: generative pre-trained transformer for biomedical text generation and mining,'' \emph{Briefings in Bioinformatics}, vol.~23, no.~6, p. bbac409, 2022.

\bibitem{chintagunta2021medically}
B.~Chintagunta, N.~Katariya, X.~Amatriain, and A.~Kannan, ``Medically aware gpt-3 as a data generator for medical dialogue summarization,'' in \emph{Machine Learning for Healthcare Conference}.\hskip 1em plus 0.5em minus 0.4em\relax PMLR, 2021, pp. 354--372.

\bibitem{li2022tod4ir}
C.~Li, X.~Zhang, D.~Chrysostomou, and H.~Yang, ``Tod4ir: A humanised task-oriented dialogue system for industrial robots,'' \emph{IEEE Access}, vol.~10, pp. 91\,631--91\,649, 2022.

\bibitem{lewis2020bart}
M.~Lewis, Y.~Liu, N.~Goyal, M.~Ghazvininejad, A.~Mohamed, O.~Levy, V.~Stoyanov, and L.~Zettlemoyer, ``Bart: Denoising sequence-to-sequence pre-training for natural language generation, translation, and comprehension,'' in \emph{Proceedings of the 58th Annual Meeting of the Association for Computational Linguistics}, 2020, pp. 7871--7880.

\bibitem{ye2023improved}
Y.~Ye, H.~You, and J.~Du, ``Improved trust in human-robot collaboration with chatgpt,'' \emph{IEEE Access}, 2023.

\bibitem{huang2023grounded}
W.~Huang, F.~Xia, D.~Shah, D.~Driess, A.~Zeng, Y.~Lu, P.~Florence, I.~Mordatch, S.~Levine, K.~Hausman \emph{et~al.}, ``Grounded decoding: Guiding text generation with grounded models for robot control,'' \emph{arXiv preprint arXiv:2303.00855}, 2023.

\bibitem{vemprala2023chatgpt}
S.~Vemprala, R.~Bonatti, A.~Bucker, and A.~Kapoor, ``Chatgpt for robotics: Design principles and model abilities,'' \emph{Microsoft Auton. Syst. Robot. Res}, vol.~2, p.~20, 2023.

\bibitem{lekova2022making}
A.~Lekova, P.~Tsvetkova, T.~Tanev, P.~Mitrouchev, and S.~Kostova, ``Making humanoid robots teaching assistants by using natural language processing (nlp) cloud-based services,'' \emph{Journal of Mechatronics and Artificial Intelligence in Engineering}, vol.~3, no.~1, pp. 30--39, 2022.

\bibitem{xie2023chatgpt}
B.~Xie, X.~Xi, X.~Zhao, Y.~Wang, W.~Song, J.~Gu, and S.~Zhu, ``Chatgpt for robotics: A new approach to human-robot interaction and task planning,'' in \emph{International Conference on Intelligent Robotics and Applications}.\hskip 1em plus 0.5em minus 0.4em\relax Springer, 2023, pp. 365--376.

\bibitem{bertacchini2023social}
F.~Bertacchini, F.~Demarco, C.~Scuro, P.~Pantano, and E.~Bilotta, ``A social robot connected with chatgpt to improve cognitive functioning in asd subjects,'' \emph{Frontiers in Psychology}, vol.~14, 2023.

\bibitem{sap2019social}
M.~Sap, H.~Rashkin, D.~Chen, R.~Le~Bras, and Y.~Choi, ``Social iqa: Commonsense reasoning about social interactions,'' in \emph{Proceedings of the 2019 Conference on Empirical Methods in Natural Language Processing and the 9th International Joint Conference on Natural Language Processing (EMNLP-IJCNLP)}, 2019, pp. 4463--4473.

\bibitem{tfidf}
\BIBentryALTinterwordspacing
scikit learn. Tfidfvectorizer. [Online]. Available: \url{https://scikit-learn.org/stable/modules/generated/sklearn.feature_extraction.text.TfidfVectorizer.html}
\BIBentrySTDinterwordspacing

\bibitem{zhang2019bertscore}
T.~Zhang, V.~Kishore, F.~Wu, K.~Q. Weinberger, and Y.~Artzi, ``Bertscore: Evaluating text generation with bert,'' in \emph{International Conference on Learning Representations}, 2019.

\bibitem{papineni2002bleu}
K.~Papineni, S.~Roukos, T.~Ward, and W.-J. Zhu, ``Bleu: a method for automatic evaluation of machine translation,'' in \emph{Proceedings of the 40th annual meeting of the Association for Computational Linguistics}, 2002, pp. 311--318.

\bibitem{bertscore}
\BIBentryALTinterwordspacing
T.~Tianyi. Bertscore. [Online]. Available: \url{https://github.com/Tiiiger/bert_score}
\BIBentrySTDinterwordspacing

\bibitem{hart2006nasa}
S.~G. Hart, ``Nasa-task load index (nasa-tlx); 20 years later,'' in \emph{Proceedings of the human factors and ergonomics society annual meeting}, vol.~50, no.~9.\hskip 1em plus 0.5em minus 0.4em\relax Sage publications Sage CA: Los Angeles, CA, 2006, pp. 904--908.

\bibitem{bartneck2009measurement}
C.~Bartneck, D.~Kuli{\'c}, E.~Croft, and S.~Zoghbi, ``Measurement instruments for the anthropomorphism, animacy, likeability, perceived intelligence, and perceived safety of robots,'' \emph{International journal of social robotics}, vol.~1, pp. 71--81, 2009.

\end{thebibliography}

\end{document}